\documentclass[lettersize,journal]{IEEEtran}
\usepackage{graphicx} 
\usepackage{subfigure}
\usepackage{makecell}
\usepackage{threeparttable}
\usepackage{multirow}
\usepackage{amsthm,amsmath,amssymb}
\usepackage{mathrsfs}
\usepackage{cite} 
\usepackage{bbding}
\usepackage{arydshln}
\usepackage{amssymb}
\DeclareMathAlphabet{\mathcal}{OMS}{cmsy}{m}{n}
\usepackage{pifont}
\usepackage{float}
\usepackage{bm}
\usepackage[colorlinks,linkcolor=blue,citecolor=black]{hyperref}
\usepackage{array}
\usepackage{booktabs}
\usepackage{soul}
\usepackage{color}
\newcommand{\tabincell}[2]{\begin{tabular}{@{}#1@{}}#2\end{tabular}}
\newcommand{\etal}{\textit{et al}. }
\newcommand{\etall}{\textit{et al}.}
\newcommand{\ie}{\textit{i}.\textit{e}.}
\newcommand{\eg}{\textit{e}.\textit{g}.}
\begin{document}

\title{A Systematic Evaluation and Benchmark for Embedding-Aware Generative Models: Features, Models, and Any-shot Scenarios}

\author{Liangjun~Feng\IEEEauthorrefmark{1},
        Jiancheng~Zhao\IEEEauthorrefmark{1},
        Chunhui~Zhao,~\IEEEmembership{Senior Member,~IEEE}

\IEEEcompsocitemizethanks{
\IEEEcompsocthanksitem This work has been submitted to the IEEE for possible publication. Copyright may be transferred without notice, after which this version may no longer be accessible.
\IEEEcompsocthanksitem This work is supported by the National Natural Science Foundation of China (No. 62125306). (Corresponding author: Chunhui Zhao.)
\IEEEcompsocthanksitem Liangjun Feng, Jiancheng Zhao and Chunhui Zhao are with College of Control Science and Engineering, Zhejiang University, Hangzhou 310027, China. (e-mail: chhzhao@zju.edu.cn; liangjunfeng@zju.edu.cn; zhaojiancheng@zju.edu.cn).
\IEEEcompsocthanksitem \IEEEauthorrefmark{1} These authors contributed equally to this work.
}
}

\markboth{ }%
{Shell \MakeLowercase{\textit{et al.}}: A Systematic Evaluation and Benchmark for Embedding-Aware Generative Models: Features, Models, and Any-shot Scenarios}
\maketitle

\begin{abstract}
Embedding-aware generative model (EAGM) addresses the data insufficiency problem for zero-shot learning (ZSL) by constructing a generator between semantic and visual feature spaces. Thanks to the predefined benchmark and protocols, the number of proposed EAGMs for ZSL is increasing rapidly. We argue that it is time to take a step back and reconsider the embedding-aware generative paradigm. The main work of this paper is two-fold. First, the embedding features in benchmark datasets are somehow overlooked, which potentially limits the performance of EAGMs, while most researchers focus on how to improve EAGMs. Therefore, we conduct a systematic evaluation of ten representative EAGMs and prove that even embarrassedly simple modifications on the embedding features can improve the performance of EAGMs for ZSL remarkably. So it's time to pay more attention to the current embedding features in benchmark datasets. Second, based on five benchmark datasets, each with six any-shot learning scenarios, we systematically compare the performance of ten typical EAGMs for the first time, and we give a strong baseline for zero-shot learning (ZSL) and few-shot learning (FSL). Meanwhile, a comprehensive generative model repository, namely, generative any-shot learning (GASL) repository, is provided, which contains the models, features, parameters, and scenarios of EAGMs for ZSL and FSL. Any results in this paper can be readily reproduced with only one command line based on GASL. 
\end{abstract}


\section{Introduction}\label{sec:introduction}
\IEEEPARstart{S}{upported} by large-scale annotated datasets \cite{1,2},  intelligent models, \eg, convolutional neural networks (CNNs)  \cite{3,4,5,6} and transformers  \cite{7,8}, have presented encouraging breakthroughs in visual recognition in the last decade. From 2012 to 2022, the top-5 accuracy of image recognition task on ImageNet1k has rapidly increased from 83\% to 98\% \cite{9,10}. However, due to the long-tailed distribution of categories, instance collection for rare objects in standard supervised learning is practically expensive and time-consuming \cite{10,11,12}. With few training exemplars or category-unbalanced datasets, popular intelligent models usually fail to present state-of-the-art results.\cite{13,14}\par   
To address the challenge, zero-shot learning (ZSL) \cite{15,16} is designed to train a model capable of recognizing unseen objects based on the dataset of seen objects. In comparison with traditional supervised learning task, there are unique techniques for ZSL to achieve knowledge transfer from seen classes to unseen classes \cite{17,18,19,20}. On the one hand, some auxiliary information, \eg, attribute or textual description of class label, is used in ZSL to bridge the class gap between seen and unseen classes. This technique, known as \textit{semantic features} \cite{21,22,23,24}, allows to identify a new object by having only a description of it. The semantic features can be obtained by manual annotation or individual representation models. For example, eighty-five manually annotated attributes are used in the benchmark Animals with Attributes (AWA) dataset \cite{15} to describe fifty kinds of animals, and 1,024-dimensional features extracted by the character-based CNN-RNN model \cite{25} are used in the benchmark Oxford Flowers (FLO) dataset \cite{26} to describe 102 kinds of flowers. On the other hand, empirical studies \cite{27,28,29,30} have indicated that replacing the original images with the features from a pre-trained CNN, known as \textit{visual features}, makes ZSL more effective for recognition task. Currently, the visual features used in benchmark datasets are 2,048-dim top-layer pooling units of ResNet101 \cite{13}, which is pre-trained on ImageNet 1K \cite{1} and not finetuned. With the well-designed semantic and visual features, models of ZSL \cite{31,32,33,34} learn from seen objects and predict for unseen objects. Particularly, when all classes are allowed at the test phase, the problem is defined as generalized zero-shot learning (GZSL) \cite{27,28,35,36,37}. In comparison with the conventional ZSL, GZSL removes the unreasonable restriction that test data only come from unseen classes and hence is more practical.\par  
Due to the importance of ZSL and GZSL, the number of new ZSL methods proposed every year has been increasing rapidly. These methods can be categorized into various paradigms, such as probability-based methods \cite{15, 17, 38},  compatibility-based methods \cite{18, 39,40,41,42}, image-aware generative models (IAGMs) \cite{43, 44, 45, 46,47}, and \textit{embedding-aware generative models} (EAGMs) \cite{48,49,50,51,52,53,54,55}. The probability-based methods, \eg, DAP \cite{15} and TIZSL \cite{38}, learn a number of probabilistic classifiers for attributes and combine the probability outputs to make predictions. The compatibility-based methods, \eg, DeViSE \cite{41} and ESZSL \cite{18}, learn a compatibility function between the semantic and visual features, which gives ranking scores for classification. As for generative models, they train conditional generators, \eg, generative adversarial networks (GANs) \cite{56, 57} and variational autoencoders (VAEs) \cite{58, 59}, based on the semantic features to synthesize virtual exemplars for unseen classes. Specifically, IAGMs aim to synthesize original images, while EAGMs aim to synthesize visual features. In comparison with other paradigms, EAGMs address the essential data insufficiency problem and enjoy both semantic features and visual features \cite{48, 50}. Hence, EAGMs have presented state-of-the-art results for ZSL and GZSL and have become more and more popular in recent years \cite{60,61,62,63,64}. In this paper, we focus on EAGMs.\par 

Particularly, we give an interesting observation that the usage of benchmark embedding features has both pros and cons for EAGMs. For clarification, we visualize EAGM for ZSL and GZSL in Figure 1. For GZSL, the paradigm of EAGM is applied as 1-2-3.a-4.a; For ZSL, EAGM is applied as 1-2-3.b-4.b. As shown, the generator $G$ is assigned to learn the mapping from semantic feature space to visual feature space. On the one hand, the embedding features contributed by Xian \etal \cite{27, 28} indeed provide a fair experimental setting. Based on the benchmark, numerous regularizations, \eg, cycle loss \cite{65, 66}, and implementations, \eg, Wasserstein GAN (WGAN) \cite{48, 56}, are proposed for $G$ to enhance its generation capability. On the other hand, we argue that the research of EAGMs on ZSL and GZSL should not be limited to the generator $G$. The embedding features deserve more attention. Essentially, ZSL and GZSL are classification tasks whose performance are upper bounded by the features. A perfect generator would make few effects with poor embedding features. In the previous works of Xian \etal \cite{67} and Narayan \etal \cite{68}, the original ResNet visual features in benchmark datasets were replaced with finetuned ResNet visual features, resulting in a significant performance improvement. Before finetuning, f-VAEGAN-D2 achieved a harmonic mean of 64.6\% on FLO for GZSL, while the number increased to 75.1\% afterward. The simple but effective trial reveals that the enhancement for embedding features is promising. However, Xian \etal \cite{67} and Narayan \etal \cite{68} did not discuss or analyze this phenomenon. Here, we take a further step to explore the effects of features for EAGMs in depth by systematic evaluation, including the visual features and semantic features.\par  

\begin{figure}[htb]
\centering
\includegraphics[width=.5\textwidth]{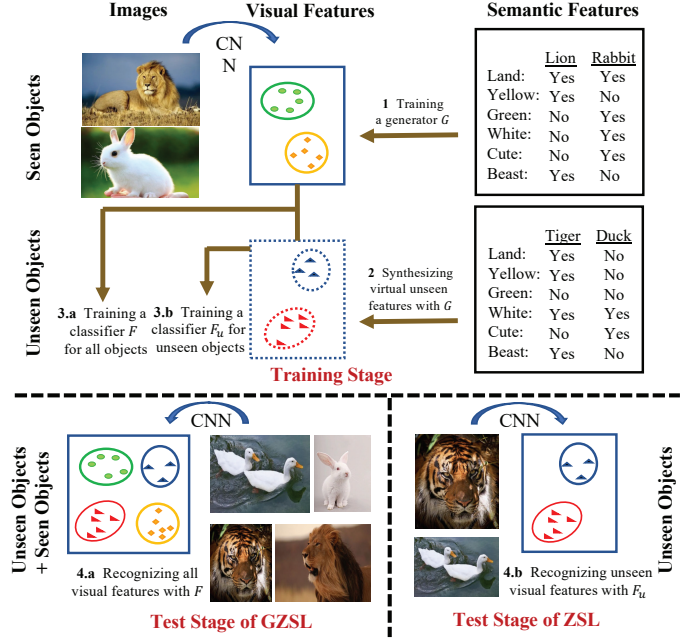} 
\caption{A schematic diagram of EAGM for ZSL and GZSL. The feature generator $G$ is constructed between the semantic and visual feature spaces. Order of GZSL:1-2-3.a-4.a; Order of ZSL:1-2-3.b-4.b.}
\end{figure}

Besides, EAGMs can be readily applied for few-shot learning (FSL) tasks \cite{70,71,72,73,74,75}. In comparison with ZSL, FSL allows a few samples per-unseen class for model training. Sometimes, the scenario where only a few samples per-seen class are used for model training is also called few-shot learning \cite{76}. To avoid misunderstanding, we name the former scenario as unseen-class few-shot learning (UFSL) and the latter scenario as seen-class few-shot learning (SFSL). Like GZSL, the more practical generalized unseen-class few-shot learning (GUFSL) and generalized seen-class few-shot learning (GSFSL) can be obtained from UFSL and SFSL respectively by making all classes available for test. The mathematical formulations for any-shot learning, \ie, ZSL, GZSL, UFSL, GUFSL, SFSL, and GSFSL, would be given in the preliminary section for clarification. Generally, the challenge of FSL is that directly training the model in the case of a few samples results in serious overfitting and unsatisfactory performance \cite{77, 78}. Because EAGMs are designed to synthesize virtual exemplars for objects, they could treat FSL as a data insufficiency problem and address it like in ZSL. Some EAGMs, \eg, CADA-VAE \cite{79} and TGG \cite{80}, have already been applied for FSL and obtained superior performance. However, distinguished from the comparison in ZSL, we argue that the comparison of EAGMs in FSL has no agreed-upon protocols. Various experimental settings may be used by different methods, leading to unfair comparisons. For example, the few samples used in model training determine the learning quality of EAGMs for FSL. Differences in the chosen samples would affect the comparison results. Besides, as shown in the FSL experiments presented by Zhang \etal \cite{80} and Schonfeld \etal \cite{79}, only one or two methods were used for comparison, which seems insufficient. This phenomenon reveals the lack of baseline results for EAGMs in FSL. It is desired to obtain a unified experimental setting and baseline for FSL like those for ZSL.\par

\begin{table*}[!htb]
  \centering
      \textbf{\caption{Summary of Six Tasks for EAGMs}}
      \renewcommand\arraystretch{1.3}
      \setlength{\tabcolsep}{1.3 mm}
      \vspace{-0.7em}
      \centering
      \begin{tabular}{cll|cccccccc}
          \Xhline{1pt}
          \multirow{2}*{\textbf{NO.}} & \multirow{2}*{\textbf{Task}} & \multirow{2}*{\textbf{Full Name}} & \multirow{2}*{\textbf{Type}}  & \multirow{2}*{\# \textbf{Methods}} & \multicolumn{2}{c}{\textbf{Training} $\rightarrow$ \textbf{Test}} & \multirow{2}*{\textbf{Challenge}}   \\
          \cline{6-7}    
           &    &   &  &  &   \textbf{Training set}  &  \textbf{Test set} &     \\
          \hline
           \textbf{1} & \textbf{ZSL}       &  \textbf{Z}ero-\textbf{S}hot \textbf{L}earning                                  &  Z  & Numerous & $\mathcal{X}_{s} \cup \mathcal{Y}_{s} \cup \mathcal{A} $ & $\mathcal{X}_{u}$ &  \multirow{2}*{Middle} \\
           \textbf{2} & \textbf{GZSL}    &  \textbf{G}eneralized \textbf{Z}ero-\textbf{S}hot \textbf{L}earning & Z & Numerous & $\mathcal{X}_{s}^{tr} \cup \mathcal{Y}_{s}^{tr} \cup \mathcal{A}$  & $\mathcal{X}_{u} \cup \mathcal{X}_{s}^{te}$ & \\ 
           \textbf{3} & \textbf{UFSL}    &  \textbf{U}nseen-class \textbf{F}ew-\textbf{S}hot \textbf{L}earning   & F  & Some    & $\mathcal{X}_{s} \cup \mathcal{Y}_{s} \cup \mathcal{X}_{u}^{tr-f} \cup \mathcal{Y}_{u}^{tr-f} \cup \mathcal{A}$  & $\mathcal{X}_{u}^{te}$ & \multirow{2}*{Low} \\
          \textbf{4}  & \textbf{GUFSL} &  \textbf{G}eneralized \textbf{U}nseen-class \textbf{F}ew-\textbf{S}hot \textbf{L}earning    & F   & Some    & $\mathcal{X}_{s}^{tr} \cup \mathcal{Y}_{s}^{tr} \cup \mathcal{X}_{u}^{tr-f} \cup \mathcal{Y}_{u}^{tr-f} \cup \mathcal{A}$    & $\mathcal{X}_{u}^{te} \cup \mathcal{X}_{s}^{te}$ & \\
          \textbf{5}  & \textbf{SFSL}    & \textbf{S}een-class \textbf{F}ew-\textbf{S}hot \textbf{L}earning   & Z+F & A Few     & $\mathcal{X}_{s}^{tr-f} \cup \mathcal{Y}_{s}^{tr-f} \cup \mathcal{A}$ & $\mathcal{X}_{u}$ & \multirow{2}*{High} \\
          \textbf{6}  & \textbf{GSFSL} &  \textbf{G}eneralized \textbf{S}een-class \textbf{F}ew-\textbf{S}hot \textbf{L}earning   & Z+F &  A Few    & $\mathcal{X}_{s}^{tr-f} \cup \mathcal{Y}_{s}^{tr-f} \cup \mathcal{A}$  & $\mathcal{X}_{u}  \cup \mathcal{X}_{s}^{te}$ & \\
          \Xhline{1pt}
          \multicolumn{8}{l}{Note: ``Z" denotes the task is a kind of zero-shot learning and ``F" denotes the task is a kind of few-shot learning.} \\
          \end{tabular}
      \vspace{-1em}
  \end{table*}  

The main contributions are as follows:
\begin{enumerate}
\item{We argue that the current features used in benchmark datasets for ZSL are somehow out-of-date, which are based on the methods proposed in 2015\cite{13} and 2016\cite{25}. For the first time, we propose to emphasize the importance of improving features for EAGMs by performing a systematic evaluation, in order to explore the potential of embedding-aware generative paradigms in depth from the perspective of visual and semantic features. It is found that even embarrassedly simple modifications on the embedding features can improve the performance of EAGMs for any-shot learning remarkably.}
\item{Different from previous reviews which usually referred to the results directly from original papers, we comprehensively reproduce and compare ten typical EAGMs based on five benchmark datasets, each with six any-shot scenarios. This work presents strong baseline results for ZSL and FSL, including ZSL, GZSL, UFSL, GUFSL, SFSL, and GSFLS, which can help future EAGMs to be sufficiently evaluated on any-shot learning. Meanwhile, we make our work a publicly available repository, namely the generative any-shot learning (GASL) repository, which contains the models, features, parameters, and settings of EAGMs for ZSL and FSL. Any results in this paper can be readily reproduced with only one command line based on GASL. Therefore, we argue that our work plays an important role in pushing EAGMs a step forward to any-shot learning.}
\end{enumerate}

This paper contains seven sections. Section II gives some preliminary works, including the basic notations and formulations. Section III reviews related works on zero-shot learning and embedding-aware generative models. Section IV focuses on the analysis of reproduced models. Section V shows our modifications on the visual features and semantic features. Section VI presents ZSL and FSL experiments for EAGMs based on benchmark datasets. A discussion of the trends for future research, along with our conclusions, is presented in Section VII.

\section{Preliminary}\label{sec:preliminary}
\subsection{Basic Notations}
For visual recognition, we have a set of seen classes, $\mathcal{Y}_{s} = \{1,...,p\}$, and a set of unseen classes, $\mathcal{Y}_{u} = \{p+1,...,p+q\}$. The two sets are disjoint, \ie, $\mathcal{Y}_{s} \cap \mathcal{Y}_{u} = \varnothing$, $p$ is the number of seen classes, and $q$ is the number of unseen classes. In comparison, seen classes have sufficient samples for model training, while unseen classes have no or few samples for model training. Suppose that a dataset of seen classes is $\mathcal{D}_{s} =  \{(x_{s}^{(i)},y_{s}^{(i)},a_{s}^{(i)}),i=1,...,N_{s}\}$, where $x_{s}^{(i)} \in \mathcal{X}_{s}$ is a visual feature with a corresponding class label $y_{s}^{(i)} \in \mathcal{Y}_{s}$ and a semantic description $a_{s}^{(i)} \in \mathcal{A}_{s}$. Similarly, a dataset of unseen classes is $\mathcal{D}_{u} =  \{(x_{u}^{(i)},y_{u}^{(i)},a_{u}^{(i)}),i=1,...,N_{u}\}$, where $x_{u}^{(i)} \in \mathcal{X}_{u}$, $y_{u}^{(i)} \in \mathcal{Y}_{u}$, and $a_{u}^{(i)} \in \mathcal{A}_{u}$. We also use $x_{s_{i}}, y_{s_{i}}, a_{s_{i}}$ and $x_{u_{i}}, y_{u_{i}}, a_{u_{i}}$ to denote the items of the $i$-th seen and unseen class, respectively. \par
Note that the visual feature $x$ is a feature of the image, $v$, \ie, $x = E_{x}(v)$, where $ E_{x}$ is the visual embedding function, \eg, the pre-trained CNN, for $v$. The semantic feature $a$ is an embedding of the class label, $y$, \ie, $a = E_{a}(t)$, where $t$ is the attribute or textual description of $y$ and $E_{a}$ is the semantic embedding function, \eg,  the character-based CNN-RNN \cite{25}, for $t$. All samples in class $y$ enjoy the same $a$. In comparison with conventional supervised learning tasks, the semantic information $a$ is introduced here as the prior knowledge of each class for model designing. 
\subsection{Problem Formulations}
In this paper, six scenarios are considered for EAGMs, including ZSL, GZSL, UFSL, GUFSL, SFSL, and GSFSL. Based on the defined notations, these tasks are formulated and explained as follows. \par
\begin{itemize}
\item[1)]  ZSL: Zero-shot learning aims to train a discriminant function based on the seen dataset and all the semantic descriptions. The evaluation of the discriminant function is performed on the unseen classes. Hence, 
the paradigm of ZSL is $\mathcal{X}_{s} \cup \mathcal{Y}_{s} \cup \mathcal{A} \rightarrow \mathcal{X}_{u}$.
\item[2)] GZSL: Generalized zero-shot learning is an extension of ZSL. Both seen and unseen classes are allowed for test. Hence, the paradigm of GZSL is $\mathcal{X}_{s}^{tr} \cup \mathcal{Y}_{s}^{tr} \cup \mathcal{A}  \rightarrow \mathcal{X}_{u} \cup \mathcal{X}_{s}^{te}$,  where $\mathcal{X}_{s}^{tr} \cup \mathcal{Y}_{s}^{tr}$ and $\mathcal{X}_{s}^{te}$ denote the training sample pairs and test samples of seen classes, respectively.
\item[3)] UFSL: Unseen-class few shot learning trains a discriminant function with the seen dataset and a few unseen sample pairs. The evaluation is performed on the unseen classes. Hence, the paradigm of UFSL is $\mathcal{X}_{s} \cup \mathcal{Y}_{s} \cup \mathcal{X}_{u}^{tr-f} \cup \mathcal{Y}_{u}^{tr-f} \cup \mathcal{A} \rightarrow \mathcal{X}_{u}^{te}$, where $\mathcal{X}_{u}^{tr-f} \cup \mathcal{Y}_{u}^{tr-f}$ denotes a few training sample pairs of unseen classes.
\item[4)] GUFSL: Generalized unseen-class few-shot learning is an extension of UFSL. Both seen and unseen classes are allowed for test. Hence, the paradigm of GUFSL is $\mathcal{X}_{s}^{tr} \cup \mathcal{Y}_{s}^{tr} \cup \mathcal{X}_{u}^{tr-f} \cup \mathcal{Y}_{u}^{tr-f} \cup \mathcal{A} \rightarrow \mathcal{X}_{u}^{te}  \cup \mathcal{X}_{s}^{te}$.
\item[5)] SFSL: Seen-class few-shot learning trains a discriminant function with only a few seen sample pairs. The evaluation is performed on the unseen classes. Hence, the paradigm of SFSL is $\mathcal{X}_{s}^{tr-f} \cup \mathcal{Y}_{s}^{tr-f} \cup \mathcal{A} \rightarrow \mathcal{X}_{u}$, where $\mathcal{X}_{s}^{tr-f} \cup \mathcal{Y}_{s}^{tr-f}$ is  a few training sample pairs of seen classes.
\item[6)] GSFSL: Generalized seen-class few-shot learning is an extension of SFSL. Both seen and unseen classes are allowed for test. Hence, the paradigm of GSFSL is $\mathcal{X}_{s}^{tr-f} \cup \mathcal{Y}_{s}^{tr-f} \cup \mathcal{A} \rightarrow  \mathcal{X}_{u} \cup \mathcal{X}_{s}^{te} $.\par
\end{itemize}

For clarification, the six tasks are listed and compared in Table I. Generally, SFSL and GSFSL can be treated as hybrid tasks of ZSL and FSL and are more challenging than others, since they have only a few seen sample pairs for model training. However, only a few EAGM-based methods \cite{76} have been proposed for SFSL and GSFSL. Most of the EAGM-based methods \cite{50,51,52,53,54,55,79,80} are designed for ZSL and GZSL, and some of them \cite{79,80} are used for UFSL and GUFSL.

\subsection{Formulations of EAGMs}
\subsubsection{Formulation of EAGMs for ZSL}
Generally, EAGMs first train a generator, $G$, based on seen classes by:
\begin{equation}
\frac{1}{N_{s}}\sum^{N_{s}}_{n=1} \mathcal{L}_{g}(x_{s},y_{s},a;W_{g}) + \Omega(W_{g}),
\end{equation}
where $\mathcal{L}_{g}$ is the loss function of $G$, $\Omega$ is the regularization term, and $W_{g}$ is the model parameter.\par
Based on the well-trained $G$, a classifier of unseen classes, $F_{u}$, can be obtained for ZSL by:
\begin{equation}
\begin{aligned}
\frac{1}{\tilde{N}_{u}}\sum^{\tilde{N}_{u}}_{n=1}& \mathcal{L}_{f_{u}}(\tilde{x}_{u},\tilde{y}_{u}, a_{u}; W_{f_{u}}), \\
&\tilde{x}_{u} = G(a_{u}, z),
\end{aligned}
\end{equation}
where $\mathcal{L}_{f_{u}}$ is the loss function of $F_{u}$, $\tilde{x}_{u}$ and $\tilde{y}_{u}$ are generated sample pairs for unseen classes, $\tilde{N}_{u}$ is the number of generated unseen samples, $W_{f_{u}}$ is the model parameter, and $z \sim \mathcal{N}(0,1)$ is the Gaussian noise. At the test stage of ZSL, the classification for unseen classes can then be readily performed by $F_{u}$.\par
For GZSL, a classifier for all classes, $F$, is obtained by:
\begin{equation}
\begin{aligned}
\frac{1}{N_{s}+\tilde{N}_{u}}\sum^{N_{s} + \tilde{N}_{u}}_{n=1}& \mathcal{L}_{f}(x_{s}, y_{s},\tilde{x}_{u},\tilde{y}_{u}, a; W_{f}),
\end{aligned}
\end{equation}
where $\mathcal{L}_{f}$ is the loss function of $F$ and $W_{f}$ is the parameter of $F$.
\subsubsection{Formulation of EAGMs for FSL}
 In UFSL and GUFSL, the generator of all classes, $G$, can be directly trained as:
\begin{equation}
\frac{1}{N_{s}+qN}\sum^{N_{s}+qN}_{n=1} \mathcal{L}_{g}(x_{s},y_{s},x_{u},y_{u},a;W_{g}) + \Omega(W_{g}),
\end{equation}
where $q$ is the number of unseen classes and $N$ is the number of provided samples for each unseen class in the training stage.\par 
For UFSL, a classifier of unseen classes, $F_{u}$, can be trained by:
\begin{equation}
\begin{aligned}
\frac{1}{qN+\tilde{N}_{u}}\sum^{qN+\tilde{N}_{u}}_{n=1}& \mathcal{L}_{f_{u}}(x_{u}, y_{u},\tilde{x}_{u} ,\tilde{y}_{u}, a_{u}; W_{f_{u}}), \\
\end{aligned}
\end{equation}
and for GUFSL, a classifier of all classes, $F$, is obtained by:
\begin{equation}
\begin{aligned}
\frac{1}{N_{s}+qN+\tilde{N}_{u}}\sum^{N_{s}+qN+\tilde{N}_{u}}_{n=1}& \mathcal{L}_{f}(x_{s}, y_{s},x_{u}, y_{u},\tilde{x}_{u} ,\tilde{y}_{u}, a; W_{f}), 
\end{aligned}
\end{equation}
where the real samples of all classes and virtual unseen samples are used in the training stage of $F$ for the final classification.\par
Besides, the formulations of zero-shot learning are applicable for SFSL and GSFSL, since SFSL and GSFSL enjoy similar data splits as ZSL and GZSL.\par

\section{Related Works}
\subsection{Review and Evaluation of ZSL and GZSL}
From the seminal paper of Lampert \etal \cite{15}, more and more works have been proposed for ZSL and GZSL. There are a few attempts have been made for literature review and method evaluation. \par
\subsubsection{Review and Evaluation of ZSL} At the early stage of ZSL, Song \etal \cite{81} reviewed about thirty methods of ZSL and categorized them into three groups based on how the feature embedding and the semantic embedding were related.  The three groups include one-order transformation models, two-order transformation models, and high-order transformation models. Also, considering the importance of semantic information for ZSL, Chen \etal \cite{82} provided a survey in the perspective of external knowledge. Four types of knowledge sources, \ie, text, attribute, knowledge, and rule, were summarized, and three paradigms, \ie, mapping function, generative model, and graph neural network, were emphasized for their promising performance on ZSL. However, this survey did not compare and analyze the performance of different patterns. Fu \etal \cite{83} reported the recent advances in ZSL from the perspective of semantic representation, model, dataset, and problem. Specifically, the semantic spaces were categorized based on whether they were attribute spaces or not. The models in ZSL were summarized in a unified ``embedding model and recognition model in embedding space'' framework.  Fifteen datasets and two problems including the projection domain shift problem and the hubness problem of ZSL were listed and discussed. From the four perspectives, although the definition and paradigm of ZSL can be introduced in detail, the performance comparison and state-of-the-art results were ignored. Similarly, in the work of Wang \etal \cite{84}, the settings, methods, and applications of ZSL were comprehensively reviewed. Based on the data utilized in model optimization, zero-shot learning methods were classified into three learning settings, including the class-inductive instance-inductive setting, class-transductive instance-inductive setting, and class-transductive instance-transductive setting. According to the recognition styles for final classification, the methods were categorized into the classifier-based methods and the instance-based methods. Representative models under each category and their applications were also introduced. These works pay more attention to the setting and category of ZSL instead of model techniques and do not analyze the generalized zero-shot learning task in depth.\par 

\begin{table*}[!htb]\scriptsize
  \centering
      \textbf{\caption{Comparison of Mentioned Reviews and Evaluations}}
      \renewcommand\arraystretch{1.3}
      \setlength{\tabcolsep}{0.7 mm}
      \centering
      \begin{tabular}{cl|lcccccccccccccccccccccc}
          \Xhline{1pt}
          \multirow{2}*{\textbf{NO.}} & \multirow{2}*{\textbf{Paper}} & & \multicolumn{6}{c}{\textbf{Task}}                      &&    \multicolumn{2}{c}{\textbf{EAGMs}}                &&    \multicolumn{2}{c}{\textbf{Embedding}}        &&    \multicolumn{4}{c}{\textbf{Sources}}                                      \\
           \cline{4-9}    \cline{11-12}     \cline{14-15}   \cline{17-20}  
                                                    &                                               &&  \textbf{ZSL} & \textbf{GZSL}  &   \textbf{UFSL} &   \textbf{GUFSL} &   \textbf{SFSL} &   \textbf{GSFSL}  &&     \textbf{Discussion}   &  \textbf{Experiments}   &&     \textbf{Discussion}   &  \textbf{Experiments}    &&  \textbf{Data/Feat.} &  \textbf{Param.} &  \textbf{Split}    &  \textbf{Code}        \\
          \hline
           \textbf{1}                           & \textbf{Song \etal \cite{81}}                 && \Checkmark &       \XSolidBrush    &     \XSolidBrush  &       \XSolidBrush    &     \XSolidBrush   &     \XSolidBrush                     &&   \XSolidBrush     &   \XSolidBrush                  &&  \Checkmark     &  \XSolidBrush       &&   \XSolidBrush    &   \XSolidBrush  &   \XSolidBrush  &   \XSolidBrush                \\
           \textbf{2}                           & \textbf{Chen \etal \cite{82}}                 && \Checkmark &       \XSolidBrush   &     \XSolidBrush    &       \XSolidBrush    &     \XSolidBrush   &     \XSolidBrush                    &&   \XSolidBrush     &   \XSolidBrush                  &&  \Checkmark     &   \XSolidBrush       &&   \XSolidBrush &   \XSolidBrush &   \XSolidBrush     &   \XSolidBrush                    \\ 
           \textbf{3}                           & \textbf{Fu \etal \cite{83}}                     &&  \Checkmark &      \XSolidBrush   &      \XSolidBrush    &       \XSolidBrush    &     \XSolidBrush   &     \XSolidBrush                   &&    \XSolidBrush      &   \XSolidBrush                 &&  \Checkmark      &  \XSolidBrush      &&   \XSolidBrush  &   \XSolidBrush  &   \XSolidBrush   &   \XSolidBrush                  \\
          \textbf{4}                           & \textbf{Wang \etal \cite{84}}                 && \Checkmark &       \XSolidBrush   &      \XSolidBrush  &       \XSolidBrush    &     \XSolidBrush   &     \XSolidBrush                              &&   \Checkmark   &    \XSolidBrush             &&   \Checkmark    &    \XSolidBrush     &&   \XSolidBrush  &   \XSolidBrush  &   \XSolidBrush &   \XSolidBrush                 \\
          \textbf{5}                            & \textbf{Xian \etal \cite{28}}                  && \Checkmark &       \Checkmark   &     \XSolidBrush    &       \XSolidBrush    &     \XSolidBrush   &     \XSolidBrush                               &&    \XSolidBrush      &   \XSolidBrush        &&   \Checkmark      &   \Checkmark      &&   \Checkmark   &   \XSolidBrush  &   \Checkmark   &   \XSolidBrush                 \\
          \textbf{6}                         & \textbf{Pourpanah \etal \cite{85}}          &&  \Checkmark &      \Checkmark &         \XSolidBrush    &       \XSolidBrush    &     \XSolidBrush   &     \XSolidBrush                              &&   \Checkmark     &    \XSolidBrush          &&   \Checkmark    &     \XSolidBrush     &&   \XSolidBrush  &   \XSolidBrush  &   \XSolidBrush     &   \XSolidBrush           \\
          \hline
          \textbf{7}                        & \textbf{Ours}                            &&  \Checkmark &    \Checkmark  &    \Checkmark  &  \Checkmark &    \Checkmark  &    \Checkmark                               && \Checkmark       & \Checkmark                     && \Checkmark       & \Checkmark              && \Checkmark     & \Checkmark  & \Checkmark     & \Checkmark               \\
          \Xhline{1pt}W
          \end{tabular}
  \end{table*}

\subsubsection{Review and Evaluation of GZSL} Xian \etal \cite{28} made a comprehensive evaluation of sixteen conventional ZSL models on the GZSL task, which revealed that conventional ZSL models suffered from the domain bias problem and showed degraded performance in GZSL. Meanwhile, Xian \etal \cite{28} provided a unified evaluation protocol and embedding features for GZSL, which contributed to transferring the research hotspot from the conventional ZSL to the more practical GZSL. However, the evaluation was still limited to the conventional ZSL models, some new paradigms, \eg, generative models, were not considered for GZSL. Pourpanah \etal \cite{85} made a comprehensive review of generalized zero-shot learning methods. More than sixty GZSL models were compared in three groups. The first group was the embedding-based methods, which learned a projection function to associate the visual features of seen classes with their corresponding semantic vectors. The second group was the generative-based methods, which learned a model to generate exemplars for unseen classes based on the seen set. The third group was the transductive-based methods, which used unlabeled unseen samples for model training. {\bf Although numerous GZSL methods were compared in this review, the results used were directly referred from previous papers. The different experimental settings in previous papers may lead to an unfair comparison of their results, which affects the conclusion in this review}.\par

For clarification, we summarize the mentioned reviews and evaluations in Table II, where the work of this paper is also listed for comparison. It can be observed that our work considers more tasks of EAGMs, and the effects of features on EAGMs are evaluated in depth and highlighted by comprehensive experiments. Different from other surveys, our work pays more attention to systematic evaluation to discuss the differences and importance of models and features. We provide a publicly available repository, which contains the models, features, parameters, and settings of EAGMs for ZSL and FSL. Based on this repository, a significant number of EAGMs can be compared in different scenarios with only one command line. Besides, there have been some reviews, surveys, and evaluations for transfer learning \cite{86,87,88,89,90} and few-shot learning \cite{91,92,93,94,95} published in recent years. Because these works do not cover the topic of embedding-aware generative models, zero-shot learning, and generalized zero-shot learning with sufficient depth, we do not discuss them in the current work. 

\subsection{Embedding-Aware Generative Models}
Embedding-aware generative models synthesize virtual visual features based on semantic descriptions, which tackle the insufficient data problem for ZSL and FSL. Thanks to superior performance and efficient training process, EAGMs have attracted more and more attention. Generally, an ideal feature generator is required to satisfy three conditions: (1) The generated virtual exemplars are similar to the real exemplars; (2) The generated virtual exemplars are discriminative for classification; (3) For ZSL, the generator trained on seen classes is semantically transferable, so that it could be applied for unseen classes. To meet the three requirements, different generative algorithms \cite{56,57}, classification regularizations \cite{48,62}, and semantic regularizations \cite{96,97,98,99,100,101} have been designed in recent years. According to the generative algorithm, we classify EAGMs into four categories, \ie, GAN-based methods, VAE-based methods, VAEGAN-based methods, and others, which are shown in Figure 2. A review of representative models of each category is then presented.\par 

\begin{figure}[htb]
\vspace{0.5em}
\centering
\includegraphics[width=.5\textwidth]{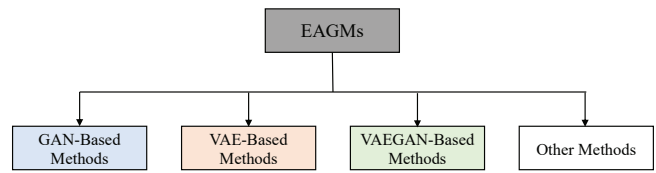} 
\caption{A generative algorithm-based classification for popular embedding-aware generative models.}
\end{figure}

\subsubsection{Review for GAN-Based Methods} GAN \cite{57} consists of a generator that synthesizes fake samples and a discriminator that distinguishes fake samples from real samples. Because the standard GAN suffers from a difficult training process, Arjovsky \etal \cite{56} applied the Wasserstein distance and constructed the Wasserstein GAN (WGAN) to improve learning stability. Based on WGAN, Xian \etal \cite{48} designed the initial feature-generating model, \ie, f-CLSWGAN, which generated discriminative features by using an external pre-trained softmax classifier. Li \etal \cite{102} introduced soul samples as the meta-representation of one class into the proposed LisGAN. By regularizing that each generated sample should be close to at least one soul sample, the model provided a more reliable generative zero-shot learning. Meanwhile, at the recognition stage, Li \etal \cite{102} proposed to use two classifiers, which were deployed in a cascade way, to achieve a coarse-to-fine result. Considering that generative models were trained using seen dataset and were expected to implicitly transfer knowledge from seen to unseen classes, Vyas \etal \cite{103} proposed the LsrGAN model, which leveraged the semantic relationship between seen and unseen categories by incorporating a novel semantic regularized loss. The regularized LsrGAN was required to generate visual features that mirrored the semantic relationships between seen and unseen classes and hence was more applicable for unseen classes.\par
\subsubsection{Review for VAE-Based Methods}  VAE \cite{59} is also a popular generative model, which uses an encoder that represents the input as a latent variable with a Gaussian distribution assumption and a decoder that reconstructs the input from the latent variable. Mishra \etal \cite{104} first used a conditional variational autoencoder (CVAE) to generate virtual samples from given attributes and applied the generated samples to train a SVM for unseen classes. In GZSL, to reduce the bias towards seen data, the SVM uses the data generated from both seen and unseen classes as opposed to using the real data for seen classes. Similarly, to address the bias problem, Schonfeld \etal \cite{79} proposed the CADA-VAE model, where a shared latent space of visual features and semantic features was learned by modality-specific aligned variational autoencoders. The key to CADA-VAE was that the distributions learned from visual features and from semantic information were aligned to construct latent features that contained the essential multi-modal information associated with unseen classes. Meanwhile, because VAEs can only optimize the lower bound on the log-likelihood of observed data, Gu \etal \cite{105} proposed a conditional version of the generative flow (GF), namely VAE-cFlow model. By using VAE, the semantic features are firstly encoded into tractable latent distributions, conditioned on that the generative flow optimizes the exact log-likelihood of the observed visual features for improved generation ability.\par
\subsubsection{Review for VAEGAN-Based Methods}  VAEGAN \cite{106}, which is a combination of VAE and GAN, is also applied for ZSL and FSL. Xian \etal \cite{67} proposed the f-VAEGAN-D2 model to combine the strengths of VAE and GAN for feature generation. In addition, via an unconditional discriminator, f-VAEGAN-D2 learns the marginal feature distribution of unlabeled samples for transductive learning. Narayan \etal \cite{107} introduced a feedback loop into VAEGAN and proposed the tf-VAEGAN model. The latent features from the feedback loop together with the corresponding synthesized features were transformed into discriminative features and utilized during classification to reduce ambiguities among categories. Similarly, Chen \etal \cite{68} proposed a simple yet effective GZSL method, termed FREE, to tackle the cross-dataset bias problem. FREE employed a feature refinement module to refine the visual features of seen and unseen classes and used a self-adaptive margin center loss to guide the model to learn semantically and class-relevant representations. An interesting application of the VAEGAN for GZSL was the GCM-CF by Yue \etal \cite{108}, which proposed a novel counterfactual faithful framework. GCM-CF applied the consistency rule to perform unseen/seen binary classification by asking whether a virtual sample of VAEGAN is counterfactual faithful or not. If a sample was classified into seen classes, the conventional supervised learning classifier could be applied; otherwise, for ZSL, the conventional ZSL classifier was applied.\par
\subsubsection{Review for Other Methods}
Apart from GANs, VAEs, and VAEGANs, a few studies have attempted to synthesize visual features with other generative algorithms. For example, Shen \etal \cite{109} incorporated the flow-based models into ZSL and proposed the invertible zero-shot flow to learn factorized data features with the forward pass of an invertible flow network. The mentioned VAE-cFlow \cite{105} can also be treated as a flow-based model. Feng \etal \cite{38} constructed a generative model with local and global relation knowledge and addressed the bias problem using a transfer increment strategy. Because these paradigms for ZSL and GZSL have not been studied by sufficient methods, we focus on GAN-, VAE-, and VAEGAN-based EAGMs in this paper.

\section{Evaluated Methods}
Here, we describe the specific models evaluated in this work. 
\vspace{-1em}
\subsection{GAN-Based Methods}
\subsubsection{f-CLSWGAN} f-CLSWGAN proposed by Xian \etal \cite{48} is the classic feature generation model, whose objective is:
\begin{equation}
\begin{aligned}
\mathcal{L}_{fclswgan} = \mathcal{L}_{wgan} + \beta\mathcal{L}_{cls},
\end{aligned}
\end{equation}
and
\begin{equation}
\begin{aligned}
\mathcal{L}_{wgan} =& \mathbb{E}[D(x_{s},a_{s})] - \mathbb{E}[D(\tilde{x}_{s},a_{s})] \\
         & + \lambda \mathbb{E}[(||\bigtriangledown_{\hat{x}_{s}}D(\hat{x}_{s}, a_{s})||_{2}-1)^{2}],
\end{aligned}
\end{equation}
\begin{equation}
\begin{aligned}
\mathcal{L}_{cls} = -\mathbb{E}[logp(y_{s}|\tilde{x}_{s})],
\end{aligned}
\end{equation}
where $\tilde{x}_{s} = G(a_{s}, z)$,  $\hat{x}_{s} = \alpha x_{s} + (1-\alpha)\tilde{x}_{s}$ with $\alpha \sim \mathcal{U}(0,1)$, $\beta$ and $\lambda$ are penalty coefficients, and $D$ and $G$ are discriminator and generator of WGAN, respectively. The $\mathcal{L}_{wgan}$ implements the adversarial training for feature generation and the $\mathcal{L}_{cls}$ makes the generated feature discriminative for classification. In f-CLSWGAN, a one-layer softmax network is used as the final classifier.\par
\subsubsection{LisGAN} LisGAN proposed by Li \etal \cite{102} introduces several soul samples for each class to guide the feature generation process. The soul sample is defined as the mean of a cluster of samples. The objective of LisGAN is:
\begin{equation}
\begin{aligned}
\mathcal{L}_{lisgan} = \mathcal{L}_{wgan} + \beta\mathcal{L}_{cls} + \delta\mathcal{L}_{r1} + \gamma\mathcal{L}_{r2},
\end{aligned}
\end{equation}
and 
\begin{equation}
\begin{aligned}
\mathcal{L}_{r1} = \frac{1}{\tilde{N}_{s}} \sum^{\tilde{N}_{s}}_{j=1, i \in \mathcal{Y}_{s}} \underset{k \in [1, K]}{min} || \tilde{x}_{s}^{(j)} -\mu^{(k)}_{s_{i}} ||^{2}_{2},
\end{aligned}
\end{equation}
\begin{equation}
\begin{aligned}
\mathcal{L}_{r2} = \frac{1}{p} \sum^{p}_{i=1} \underset{k \in [1, K]}{min} || \tilde{\mu}^{(k)}_{s_{i}} -\mu^{(k)}_{s_{i}} ||^{2}_{2},
\end{aligned}
\end{equation}
where $\tilde{N}_{s}$ is the number of generated samples, $K$ is the number of soul samples, $\mu^{(k)}_{s_{i}}$ is the $k$-th soul sample of the $i$-th seen classes, and $\delta$ and $\gamma$ are penalty coefficients.  The $\mathcal{L}_{r1}$ and $\mathcal{L}_{r2}$ require the generated sample and the virtual soul sample to be close to at least one soul sample. Besides, in the final classification stage, LisGAN uses two classifiers in a cascade style to achieve a coarse-to-fine result.
\subsubsection{LsrGAN} LsrGAN proposed by Vyas \etal  \cite{103} introduces a SR-Loss to require the generated visual features to mirror the semantic relationships among classes. The objective of LsrGAN is:
\begin{equation}
\begin{aligned}
\mathcal{L}_{lsrgan} = \mathcal{L}_{wgan} + \beta\mathcal{L}_{cls} + \delta\mathcal{L}_{sr1} +  \gamma\mathcal{L}_{sr2},
\end{aligned}
\end{equation}
and
\begin{equation}
\begin{aligned}
\mathcal{L}_{sr1} &= \frac{1}{p} \sum^{p}_{i=1} \sum_{j=1}^{p} [||max(0,C(\mu_{s_{j}}, \tilde{\mu}_{s_{i}})-(C(a_{s_{j}}, a_{s_{i}}) +\epsilon))||^{2}_{2} \\
&+ ||max(0,(C(a_{s_{j}}, a_{s_{i}}) - \epsilon)-C(\mu_{s_{j}}, \tilde{\mu}_{s_{i}})||^{2}_{2}],
\end{aligned}
\end{equation}
\begin{equation}
\begin{aligned}
\mathcal{L}_{sr2} &= \frac{1}{q} \sum^{q}_{i=1} \sum_{j=1}^{p} [||max(0,C(\mu_{s_{j}}, \tilde{\mu}_{u_{i}})-(C(a_{s_{j}}, a_{u_{i}}) +\epsilon))||^{2}_{2} \\
&+ ||max(0,(C(a_{s_{j}}, a_{u_{i}}) - \epsilon)-C(\mu_{s_{j}}, \tilde{\mu}_{u_{i}})||^{2}_{2}],
\end{aligned}
\end{equation}
where $\mu$ is the mean of a class's visual features, $C$ is the cosine similarity, and $\epsilon$ is the given error bound. The $\mathcal{L}_{sr1}$ and $\mathcal{L}_{sr2}$ are performed in the seen scope and between seen and unseen scopes, respectively, which require the visual relationships among classes to be similar to semantic relationships. The one-layer softmax classifier is also used for LsrGAN.
\subsection{VAE-Based Methods}
\subsubsection{CVAE} CVAE, which consists of an encoder, $E$, and a decoder, $G$, is first used by Mishra \etal \cite{104} to generate unseen visual features. The objective of CVAE in ZSL is:
\begin{equation}
\begin{aligned}
\mathcal{L}_{cvae} = \mathcal{L}_{kl} +  \beta \mathcal{L}_{rec},
\end{aligned}
\end{equation}
and
\begin{equation}
\begin{aligned}
\mathcal{L}_{kl} = KL(q(h_{s}|a_{s})||p(h_{s}|x_{s},a_{s})),
\end{aligned}
\end{equation}
\begin{equation}
\begin{aligned}
 \mathcal{L}_{rec}  = - \mathbb{E}[logp(x_{s}|h_{s},a_{s})],
\end{aligned}
\end{equation}
where $h_{s} = E(x_{s}, a_{s})$ is the hidden feature of CVAE. The $\mathcal{L}_{kl}$ requires $h_{s}$ to follow the given distribution and the $\mathcal{L}_{rec}$ reconstructs the input sample from the hidden feature, \ie, $\tilde{x}_{s} = G(h_{s}, a_{s})$. In CVAE, SVM is used as the final classifier. For GZSL, the data generated from both seen and unseen classes are used for training to reduce the bias towards seen classes. 
\subsubsection{CADA-VAE} CADA-VAE proposed by Schonfeld \etal \cite{79} aligns two VAEs, \ie, xVAE and aVAE, to learn a shared latent space of visual features and semantic features. The objective of CADA-VAE is:
\begin{equation}
\begin{aligned}
\mathcal{L}_{cadavae} = \mathcal{L}_{xvae} + \mathcal{L}_{avae} + \delta\mathcal{L}_{ca} + \gamma\mathcal{L}_{da},
\end{aligned}
\end{equation}
and 
\begin{equation}
\begin{aligned}
\mathcal{L}_{xvae} = KL(q(h_{sx})||p(h_{sx}|x_{s})) - \beta \mathbb{E}[logp(x_{s}|h_{sx})],
\end{aligned}
\end{equation}
\begin{equation}
\begin{aligned}
\mathcal{L}_{avae} = KL(q(h_{sa})||p(h_{sa}|a_{s})) - \beta \mathbb{E}[logp(a_{s}|h_{sa})],
\end{aligned}
\end{equation}
\begin{equation}
\begin{aligned}
\mathcal{L}_{ca} = |x_{s} - G_{x}(E_{a}(a_{s}))| + |a_{s} - G_{a}(E_{x}(x_{s}))|,
\end{aligned}
\end{equation}
\begin{equation}
\begin{aligned}
\mathcal{L}_{da} = (||\mu_{h_{sa}}-\mu_{h_{sx}}||^{2}_{2} + ||\sigma^{\frac{1}{2}}_{h_{sa}} - \sigma^{\frac{1}{2}}_{h_{sx}}||^{2}_{2})^{\frac{1}{2}},
\end{aligned}
\end{equation}
where $h$ is the hidden feature of VAE, and $\mu$ and $\sigma$ are the mean and variance of $h$. The $\mathcal{L}_{xvae}$ and $\mathcal{L}_{avae}$ train two VAEs for $x$ and $a$, respectively, the $\mathcal{L}_{ca}$ performs a cross-modal reconstruction, and the $\mathcal{L}_{da}$ aligns two hidden spaces to overcome the bias between visual and semantic features. In CADA-VAE, the one-layer softmax network is used as the final classifier.
\subsubsection{VAE-cFlow} VAE-cFlow proposed by Gu \etal \cite{105} uses the GF model to accurately estimate the log-likelihood of observed data. The objective of VAE-cFlow is:
\begin{equation}
\begin{aligned}
\mathcal{L}_{vaecflow} = \mathcal{L}_{flow} + \delta \mathcal{L}_{vae}^{flow}  + \gamma\mathcal{L}_{hcls},
\end{aligned}
\end{equation}
and
\begin{equation}
\begin{aligned}
\mathcal{L}_{flow} &= -\mathbb{E}[logp(x_{s}|a_{s})] \\
                             &= -\mathbb{E}[logp(h_{sf}|a_{s}) + log(|det(\frac{dh_{sf}}{dx_{s}})|) ],
\end{aligned}
\end{equation}
\begin{equation}
\begin{aligned}
 \mathcal{L}_{vae}^{flow} =  -\mathbb{E}[logp(a_{s}|h_{sf})] + \sum(\sigma_{h_{sf}}^{2} -log\sigma^{2}_{h_{sf}}-1),
\end{aligned}
\end{equation}
\begin{equation}
\begin{aligned}
\mathcal{L}_{hcls} = -\mathbb{E}[logp(y_{s}|h_{s})],
\end{aligned}
\end{equation}
where $h_{s}$ is the hidden feature of VAE for $a_{s}$, $h_{sf} $ is the scaled $h_{s}$. The $ \mathcal{L}_{flow}$ trains an invertible generative model between $h_{sf}$ and $x_{s}$, while the $\mathcal{L}_{vae}^{flow}$ trains a VAE model without the zero-mean restriction to encode $a_{s}$ to $h_{s}$. The $\mathcal{L}_{hcls}$ is applied on $h_{s}$ to encourage the separation of different classes. The one-layer softmax network is used for VAE-cFlow.
\subsection{VAEGAN-Based Methods}
\subsubsection{f-VAEGAN-D2} f-VAEGAN-D2 proposed by Xian \etal \cite{67} combines the strength of CVAE and WGAN. In f-VAEGAN-D2, the decoder of VAE is the generator of WGAN. The objective of f-VAEGAN-D2 is:
\begin{equation}
\begin{aligned}
\mathcal{L}_{fvaegand2} &= \mathcal{L}_{vaegan} +  \gamma \mathcal{L}^{u}_{wgan} \\
&= \mathcal{L}_{wgan} + \delta \mathcal{L}_{cvae} +  \gamma \mathcal{L}^{u}_{wgan} ,
\end{aligned}
\end{equation}
and
\begin{equation}
\begin{aligned}
\mathcal{L}^{u}_{wgan} = & \mathbb{E}[D(x_{u})] - \mathbb{E}[D(\tilde{x}_{u})] \\
         & + \lambda \mathbb{E}[(||\bigtriangledown_{\hat{x}_{u}}D(\hat{x}_{u})||_{2}-1)^{2}],
\end{aligned}
\end{equation}
where $\mathcal{L}^{u}_{wgan}$ is applied for the transductive learning of unseen classes. In this paper, we only consider the inductive learning setting, and f-VAEGAN-D2 is performed without $\mathcal{L}^{u}_{wgan}$. The one-layer softmax network is used for classification.
\subsubsection{tf-VAEGAN} tf-VAEGAN proposed by Narayan \etal \cite{107} introduces a feedback loop into VAEGAN. The objective of tf-VAEGAN is:
\begin{equation}
\begin{aligned}
\mathcal{L}_{tfvaegan} &= \mathcal{L}_{vaegan} +  \gamma \mathcal{L}_{cyc},
\end{aligned}
\end{equation}
and 
\begin{equation}
\begin{aligned}
\mathcal{L}_{cyc} = \mathbb{E}[|Dec(x_{s})-a_{s}| + |Dec(\tilde{x}_{s})-a_{s}| ],
\end{aligned}
\end{equation}
where $Dec$ is a decoder to obtain $\hat{a}_{s} = Dec(x_{s})$ and $\tilde{a}_{s} =Dec(\tilde{x}_{s})$. The $\mathcal{L}_{cyc}$ reconstructs the sample back to its semantic description. The latent feature $h_{s}$ in reconstruction is used as a part of the generator input, which enhances the virtual features' semantical consistency. In tf-VAEGAN, the one-layer softmax network is used as the final classifier.
\begin{table*}[!htb]\scriptsize
  \centering
      \textbf{\caption{Comparison of Evaluated Models}}
      \renewcommand\arraystretch{1.3}
      \setlength{\tabcolsep}{0.2 mm}
      \vspace{-0.7em}
      \centering
      \begin{tabular}{cl|cccccccccccccccccccc}
          \Xhline{1pt}
          \multirow{2}*{\textbf{NO.}} &  \multirow{2}*{\textbf{Model}}       && \multicolumn{2}{c}{\textbf{Generative Model}} 
         
                                                                                                                                        && \multicolumn{3}{c}{\textbf{Novel Regularization}}         &&    \multicolumn{2}{c}{\textbf{Classification}}     &  \multirow{2}*{\tabincell{c}{\textbf{Additional} \\ \textbf{Techniques}} }                           \\
         \cline{4-5}  \cline{7-9}         \cline{11-12} 
                               &                                             &&   \textbf{Type}  &  \textbf{Loss}    &&  \textbf{Discrim.} & \textbf{Generat.}  &   \textbf{Vis.-Sem.} &&     \textbf{Classifier}   &  \textbf{Train. Input Data}                           \\
          \hline
           \textbf{1}    &     \textbf{ f-CLSWGAN} \cite{48}            &&   WGAN  &  $\mathcal{L}_{wgan}$       && $\mathcal{L}_{cls}$ &      ---               &     ---                             &&   Softmax    &  $x_{s} \cup \tilde{x}_{u}$           &     ---                     \\
           \textbf{2}    &     \textbf{ LisGAN} \cite{102}                    &&   WGAN &  $\mathcal{L}_{wgan}$       && $\mathcal{L}_{cls}$ &   $\mathcal{L}_{r1} + \mathcal{L}_{r2}$  &     ---                               &&     Softmax     &   $x_{s} \cup \tilde{x}_{u}$   & \tabincell{c}{Two-step classification}                                 \\ 
           \textbf{3}    &     \textbf{ LsrGAN} \cite{103}                 &&   WGAN  &  $\mathcal{L}_{wgan}$       && $\mathcal{L}_{cls}$ &    ---   &     $\mathcal{L}_{sr1} +  \mathcal{L}_{sr2}$                             &&   Softmax      &    $x_{s} \cup \tilde{x}_{u}$   &  ---                \\
          \hline
          \textbf{4}     &     \textbf{CVAE} \cite{104}                       &&   CVAE  &  $\mathcal{L}_{cvae}$            && --- &      ---   &    ---                              &&   SVM   &  $\tilde{x}_{s} \cup \tilde{x}_{u}$       &  ---                                \\
          \textbf{5}     &     \textbf{CADA-VAE} \cite{79}             &&   VAE &  $\mathcal{L}_{xvae} + \mathcal{L}_{avae} $            && --- &     $\mathcal{L}_{ca}$  &    $ \mathcal{L}_{da} $                                &&    Softmax      &   $ h_{sx} \cup \tilde{h}_{ux}$     & Warm-up stage                  \\
          \textbf{6}     &     \textbf{VAE-cFlow} \cite{105}        &&  GF/VAE  & $\mathcal{L}_{flow} + \mathcal{L}_{vae}^{flow}$            &&  $\mathcal{L}_{hcls}$ &   --- &    ---                                &&   Softmax     &   $x_{s} \cup \tilde{x}_{u}$    & Temperature scaling                         \\
          \hline
          \textbf{7}     &     \textbf{f-VAEGAN-D2} \cite{67}     &&   VAEGAN  &  $\mathcal{L}_{vaegan}$          &&  --- &    $ \mathcal{L}^{u}_{wgan}$  &    ---                               && Softmax        &  $x_{s} \cup \tilde{x}_{u}$       &  ---                                   \\
          \textbf{8}     &     \textbf{tf-VAEGAN} \cite{107}          &&   VAEGAN  &  $\mathcal{L}_{vaegan}$          &&  --- &    --- &        $ \mathcal{L}_{cyc}$                                &&  Softmax    &    $x_{s} \cup h_{s}  \cup \tilde{x}_{u} \cup \tilde{h}_{u}$      & Feed-back loop                       \\
          \textbf{9}     &     \textbf{FREE}  \cite{68}                  &&   VAEGAN  &  $\mathcal{L}_{vaegan}$           && $ \mathcal{L}_{samc}$  &   ---  &   $\mathcal{L}_{cyc}$       && Softmax       &  $x_{s} \cup h_{s} \cup \hat{a}_{s}  \cup \tilde{x}_{u} \cup \tilde{h}_{u} \cup \tilde{a}_{u}$       & ---                      \\
          \textbf{10}  &      \textbf{GCM-CF} \cite{108}              &&   VAEGAN  & $\mathcal{L}_{vaegan}$           &&  $\mathcal{L}_{y}$ &    ---  &   ---                              &&  Softmax       & $x_{s} \cup h_{s}  \cup \tilde{x}_{u} \cup \tilde{h}_{u}$            &      Counterfactual inference                     \\
          \Xhline{1pt}
          \end{tabular}
  \end{table*}

\subsubsection{FREE} FREE proposed by Chen \etal \cite{68} uses a new SAMC loss to learn class- and semantically relevant representations. The objective of FREE is:
\begin{equation}
\begin{aligned}
\mathcal{L}_{free} = \mathcal{L}_{vaegan} +  \gamma \mathcal{L}_{cyc} + \xi \mathcal{L}_{samc},
\end{aligned}
\end{equation}
and
\begin{equation}
\begin{aligned}
\mathcal{L}_{samc} = max(0, \Delta + \eta||\mu_{s}-c_{y}||^{2}_{2} - (1-\eta)||\mu_{s}-c_{{y}'}||^{2}_{2}),
\end{aligned}
\end{equation}
where $\mu_{s}$ is encoded from the hidden feature $h_{s}$ in the reconstruction by $\mathcal{L}_{cyc}$, $\Delta$, $\xi$, and $\eta$ are given coefficients, $c_{y}$ is the trainable center of class $y$, and $c_{{y}'}$ is the center of another class. The $\mathcal{L}_{samc}$ makes $\mu_{s}$ intra-class contraction and inter-class separation. FREE also uses the softmax classifier for final classification. 
\subsubsection{GCM-CF} GCM-CF proposed by Yue \etal \cite{108} designs a novel counterfactual inference framework. The objective of GCM-CF is:
\begin{equation}
\begin{aligned}
\mathcal{L}_{gcmcf} = \mathcal{L}_{vaegan} +  \gamma \mathcal{L}_{y},
\end{aligned}
\end{equation}
and 
\begin{equation}
\begin{aligned}
\mathcal{L}_{y} = -log\frac{exp(-dist(x_{s},\tilde{x}_{s}))}{\sum_{{\tilde{x}}'_{s} \in {\mathcal{\tilde{X}}}'_{s} \cup \{\tilde{x}_{s}\} } exp(-dist(x_{s},{\tilde{x}}'_{s})) },
\end{aligned}
\end{equation}
where $dist$ denotes Euclidean distance and ${\tilde{x}}'_{s}$ is the generated counterfactual samples. Based on the designed counterfactual inference rules, GCM-CF performs unseen/seen binary classification for a test sample. And the tf-VAEGAN model would be applied for the final supervised or ZSL clssification.\par
For clarification, the reviewed ten models are listed and compared in Table III, where the mentioned regularizations are categorized into discriminative items (Discrim.), generative items (Generat.), and visual-semantic items (Vis.-Sem.).   
\section{Embedding Modifications}
Here, we modify the benchmark features contributed by Xian \etal \cite{28} to explore the effects of features on EAGMs in depth, including visual features and semantic features.

\subsection{Modifications on Visual Embedding}
\subsubsection{Original Features} Currently, the visual features used in benchmark datasets are 2,048-dim top-layer pooling units of ResNet101 \cite{13}, which is pre-trained on ImageNet 1K and not finetuned. The ResNet features are validated to be more effective than GoogLeNet features \cite{102} and widely used in research. We name the ResNet features contributed by Xian \etal \cite{28} as ``original features''.
\subsubsection{Naive Features} To provide a fair comparison, we reproduce the 2,048-dimensional features with a pre-trained ResNet101 network. Specifically, we resize the images in the dataset to $256 \times 256$ and then apply a center cropping function to obtain $224 \times 224$ images as the input of ResNet101. The 2,048-dim top-layer pooling units are used as the visual representation. We name the reproduced visual features as ``naive features''.
\subsubsection{Finetuned Features} We finetune the pre-trained ResNet101 on the seen classes of each dataset. Note that only the images in the training set are used and the splits proposed by Xian \etal \cite{28} are followed to make the zero-shot setting effective. The training loss is:
\begin{equation}
\begin{aligned}
\mathcal{L}_{ce} = -\mathbb{E}[logp(y_{s}|x_{s})].
\end{aligned}
\end{equation}
We observe that the SGD optimizer is more effective than the Adam optimizer \cite{113}. The momentum of SGD is set to 0.9. The learning rate is initialized to 0.01, and a linear learning rate decay scheduler is used to decrease the learning rate by 0.1 every 7 rounds for better convergence. We name the features as ``finetuned features''. In comparison with the naive features, the finetuned features contain more visual information about the benchmark dataset. 
\subsubsection{Regularized Features} We also finetune the pre-trained ResNet101 with a semantic regularization item. The training loss is:
\begin{equation}
\begin{aligned}
\mathcal{L}_{re} = \mathcal{L}_{ce} + \alpha \mathcal{L}_{se},
\end{aligned}
\end{equation}
and 
\begin{equation}
\begin{aligned}
\mathcal{L}_{se} = max(0, \Delta+\lambda C(x_{s},a_{s})-(1-\lambda)C(x_{s},{a}'_{s})),
\end{aligned}
\end{equation}
where  $\Delta$,  $\alpha$, and $\lambda$ are given coefficients, $a_{s}$ is the semantic description of $x_{s}$, ${a}'_{s}$ is the semantic description of another class, and $C$ is the cosine similarity. In comparison with conventional triplet loss, the $\mathcal{L}_{se}$ makes the visual feature $x_{s}$ align with its semantic description $a_{s}$ and misalign with other descriptions. We name the features as ``regularized features''. In comparison with the finetuned features, regularized features contain semantic information of the benchmark dataset.

\subsection{Modifications on Semantic Embedding}
\subsubsection{Original Descriptions}  The character-based CNN-RNN model \cite{25} is currently used to obtain the benchmark 1,024-dimensional semantic features. We name the semantic descriptions contributed by Xian \etal \cite{28} as ``original descriptions''. Meanwhile, we note that some datasets \cite{110, 111, 28, 15} use manually annotated attributes as semantic features. Since the modifications of manually annotated attributes depend on experience and have little algorithm contribution, we pay little attention to them in this work.

\subsubsection{Naive Descriptions} Similarly, to provide a fair comparison, we reproduce the character-based CNN-RNN model to obtain the 1,024-dimensional descriptions. The backbone of the model's image feature extractor is GoogLeNet \cite{102}, and the backbone of the model's text feature extractor is LSTM \cite{114}. We use the texts and class splits contributed by Xian \etal \cite{28}.  The symmetrical training loss of the character-based CNN-RNN model is:
\begin{equation}
\begin{aligned}
\mathcal{L}_{dssje} = \mathcal{L}_{xsje} + \mathcal{L}_{asje},
\end{aligned}
\end{equation}
and 
\begin{equation}
\begin{aligned}
\mathcal{L}_{xsje} = \underset{y \in \mathcal{Y}_{s}}{max}        (0, \Gamma(y_{s}, y) +  \underset{a \in \mathcal{A}_{s}}{\mathbb{E}}[C(x_{s},a)-C(x_{s},a_{s})] ),
\end{aligned}
\end{equation}
\begin{equation}
\begin{aligned}
\mathcal{L}_{asje} = \underset{y \in \mathcal{Y}_{s}}{max}        (0, \Gamma(y_{s}, y) +  \underset{x \in \mathcal{X}_{s}}{\mathbb{E}}[C(x,a_{s})-C(x_{s},a_{s})] ),
\end{aligned}
\end{equation}
where $\Gamma$ is the 0-1 loss and $C$ is the cosine similarity. The $\mathcal{L}_{xsje} $ and $\mathcal{L}_{asje}$ optimize the visual-semantic compactness based on $x_{s}$ and $a_{s}$, respectively. The average hidden features of LSTM are used as the semantic descriptions for each class and are named ``naive descriptions''. 
\subsubsection{GRU-based Descriptions} To follow the zero-shot principle, the character-based CNN-RNN model is trained on seen classes and applied to obtain all classes' semantic descriptions. As discussed in previous research on conventional ZSL, this paradigm suffers from a bias problem toward seen classes during predicting for unseen classes. In experiments, we notice that using GRU \cite{115} replaces LSTM as the the backbone of text feature extractor could alleviate the model's overfitting on seen classes with fewer parameters and contributes to performance improvement. We name the 1,024-dimensional features from the modified character-based CNN-GRU model as ``GRU-based descriptions''.
\subsubsection{Imbalanced GRU-based Descriptions} The character-based CNN-GRU model enjoys a balanced training loss, \ie, $\mathcal{L}_{dssje}$, which equally treats the visual features, $x_{s}$, and the semantic features, $a_{s}$, by $\mathcal{L}_{xsje}$ and $\mathcal{L}_{asje}$, respectively. However, we note that the aim of the character-based CNN-GRU for ZSL is the semantic features, and the visual features of the model are usually not used for other models. The semantic feature-based loss deserves more attention in model training. Hence, the objective of the character-based CNN-GRU is modified as:
\begin{equation}
\begin{aligned}
\mathcal{L}_{disje} = (1-\alpha) \mathcal{L}_{xsje} + \alpha \mathcal{L}_{asje},
\end{aligned}
\end{equation}
where $\alpha > 0.5$ is the given weight. We name the 1,024-dimensional features from the character-based CNN-GRU model trained by  $\mathcal{L}_{disje}$ as ``imbalanced GRU-based descriptions''.\par

\section{Experiments}
\subsection{Experiment Settings}
\subsubsection{Benchmark Dataset} In this paper, five benchmark datasets are used for experiments, including the Oxford Flowers (FLO) dataset \cite{26}, the Caltech-UCSD-Birds (CUB) dataset \cite{110}, the SUN attributes (SUN) dataset \cite{111}, the Animals with Attributes2 (AWA2) dataset \cite{28}, and the Animals with Attributes (AWA) dataset \cite{15}. FLO contains 8,189 flower images from 102 different categories. CUB is a fine-grained dataset, which contains 11,788 images from 200 kinds of birds. SUN is a subset of the SUN scene dataset with fine-grained attributes, which contains 14,340 images from 717 types of scenes. AWA2 and AWA are classic image datasets and include fifty types of animals. AWA2 has 37,322 images and AWA has 30,475 images. Note that the original images of AWA are unavailable for research, but the features of AWA are contributed by Xian \etal \cite{27}. The statistics of the five datasets are summarized in Table IV. 

\begin{table}[!htb]\scriptsize
  \centering
      \textbf{\caption{Statistics of the Five Benchmark Datasets}}
      \renewcommand\arraystretch{1.1}
      \setlength{\tabcolsep}{3 mm}
      \vspace{-0.7em}
      \centering
      \begin{tabular}{c|ccccccc}
          \Xhline{1pt}
          \multirow{2}*{\textbf{Dataset}} & \multirow{2}*{\#$\mathcal{A}$} & \multicolumn{2}{c}{\textbf{Class number}} & &  \multicolumn{3}{c}{\textbf{Image number}}  \\
          \cline{3-4}         \cline{6-8}
              &     &  \#$\mathcal{Y}_{s}$  &   \#$\mathcal{Y}_{u}$ && \textbf{Total} &  \#$\mathcal{X}_{s}$   &   \#$\mathcal{X}_{u}$ \\
          \hline
          \textbf{FLO}  & 1024& 82  & 20 && 8189  & 7034  & 1155 \\
          \textbf{CUB}  & 312 & 150 & 50 && 11788 & 8821  & 2967 \\
          \textbf{SUN}  & 102 & 645 & 72 && 14340 & 12900 & 1440 \\
          \textbf{AWA2} & 85  & 40  & 10 && 37322 & 29409 & 7913 \\
          \textbf{AWA}  & 85  & 40  & 10 && 30475 & 25517 & 4958 \\
          \Xhline{1pt}
          \end{tabular}
      \vspace{-1em}
  \end{table}

\subsubsection{Embedding features} The visual features and semantic descriptions used for experiments have been discussed in last section, and they are listed and compared in Table V for clarification. Besides, we visualize the visual features and semantic features on the unseen classes of FLO by t-SNE algorithm \cite{116} for intuitive comparison. The visualization of four types of visual features is shown in Figure 3, and the sample-level visualization of three types of semantic features is shown in Figure 4. Note that Xian \etal \cite{28} did not contribute the sample-level semantic features of original descriptions. Hence, the original descriptions are not visualized here. From Figure 3 and Figure 4, we can observe that the distributions of these features are very different, though they are obtained in a similar paradigm.

\begin{table}[!htb]\scriptsize
  \centering
      \textbf{\caption{Comparison of Visual and Semantic Emebddings}}
      \renewcommand\arraystretch{1.3}
      \setlength{\tabcolsep}{0.01 mm}
      \vspace{-0.7em}
      \centering
      \begin{tabular}{c|cccccccccccccccccccc}
          \Xhline{1pt}
           \textbf{Scope}   & \textbf{Embedding}    & \textbf{Information Source}   & \textbf{Dataset}  \\
          \hline
          \multirow{4}*{$\mathcal{X}$}         &   Original         & ImageNet                      &   FLO,CUB,SUN,AWA2,AWA         \\
                                                                 &    Naive           & ImageNet                         &   FLO,CUB,SUN,AWA2                               \\ 
                                                                 &   Finetuned     & ImageNet + Seen images                               &   FLO,CUB,SUN,AWA2      \\
                                                                 &   Reg.             & ImageNet + Seen images + Descrip.         &   FLO,CUB,SUN,AWA2           \\
           \hline
          \textbf{Scope}  & \textbf{Embedding}    & \textbf{Model}   & \textbf{Dataset}  \\
           \hline
           \multirow{4}*{$\mathcal{A}$}         &   Original        & CNN-RNN/Att.     &  FLO/CUB,SUN,AWA2,AWA            \\
                                                                &   Naive            & CNN-RNN                    &  FLO                         \\
                                                                  &   GRU.           & CNN-GRU                    &   FLO            \\
                                                             &   Imb. GRU.        & Imb. CNN-GRU           &   FLO              \\
          \Xhline{1pt}
          \multicolumn{4}{l}{Note A: Visual images of AWA are unavailable for research.} \\
          \multicolumn{4}{l}{Note B: Original features are made by Xian \etall, and others are by ourselves.} \\
          \end{tabular}
  \end{table}

\begin{figure}[!htb]
\centering 
\subfigure[Original features]{
\includegraphics[width=4cm]{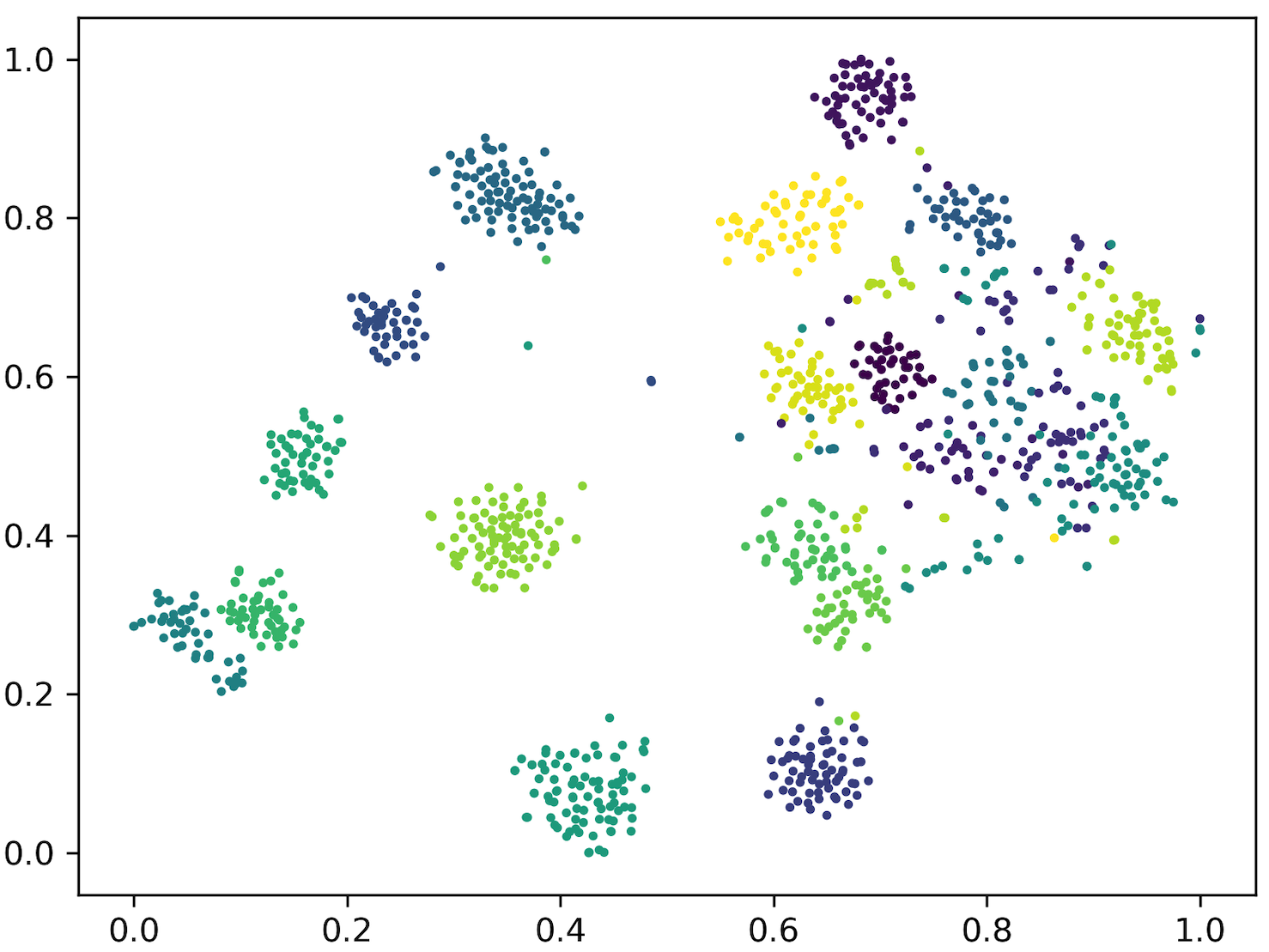}}
\subfigure[Naive features]{
\includegraphics[width=4cm]{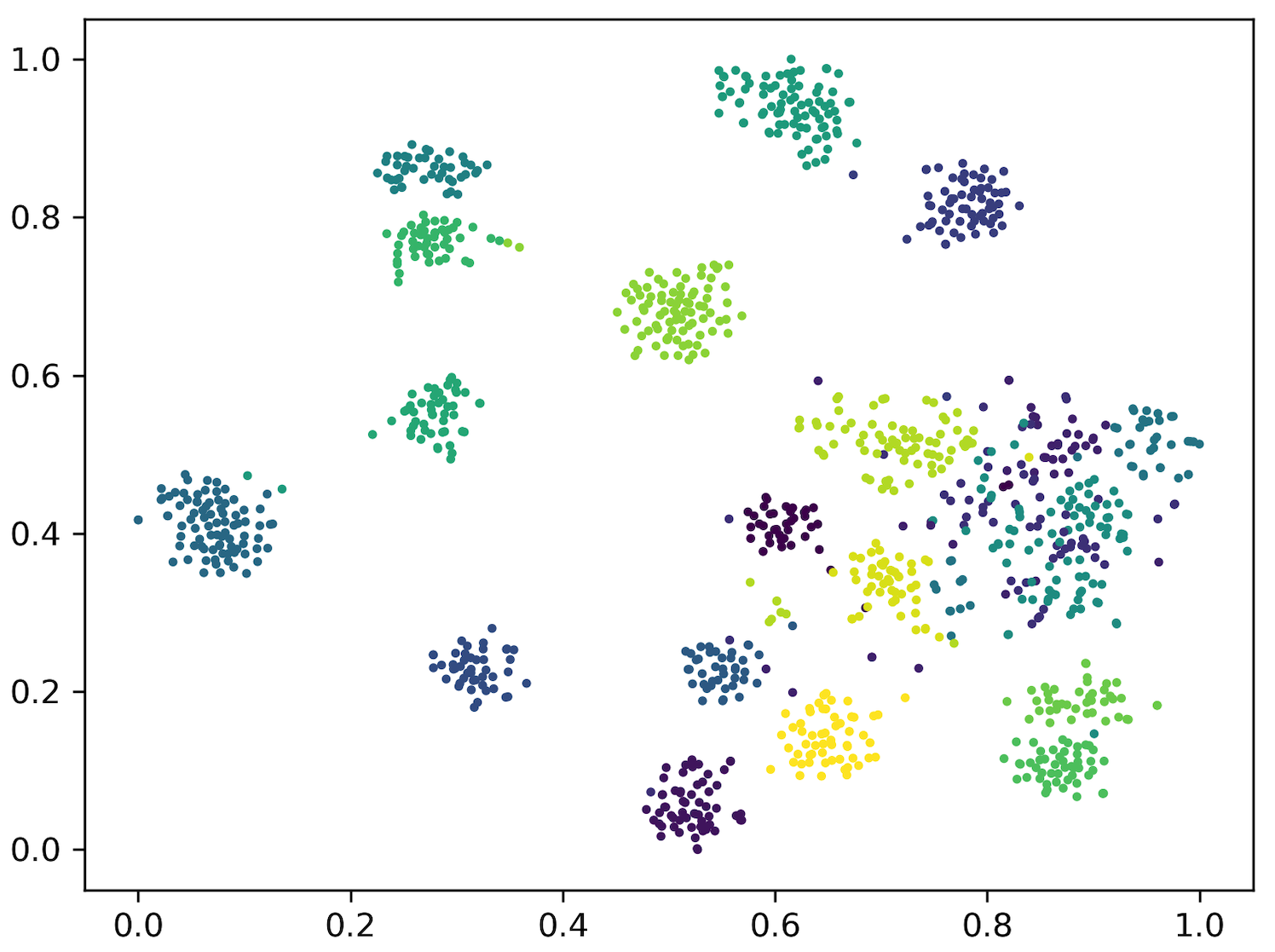}}
\subfigure[Finetuned features]{
\includegraphics[width=4cm]{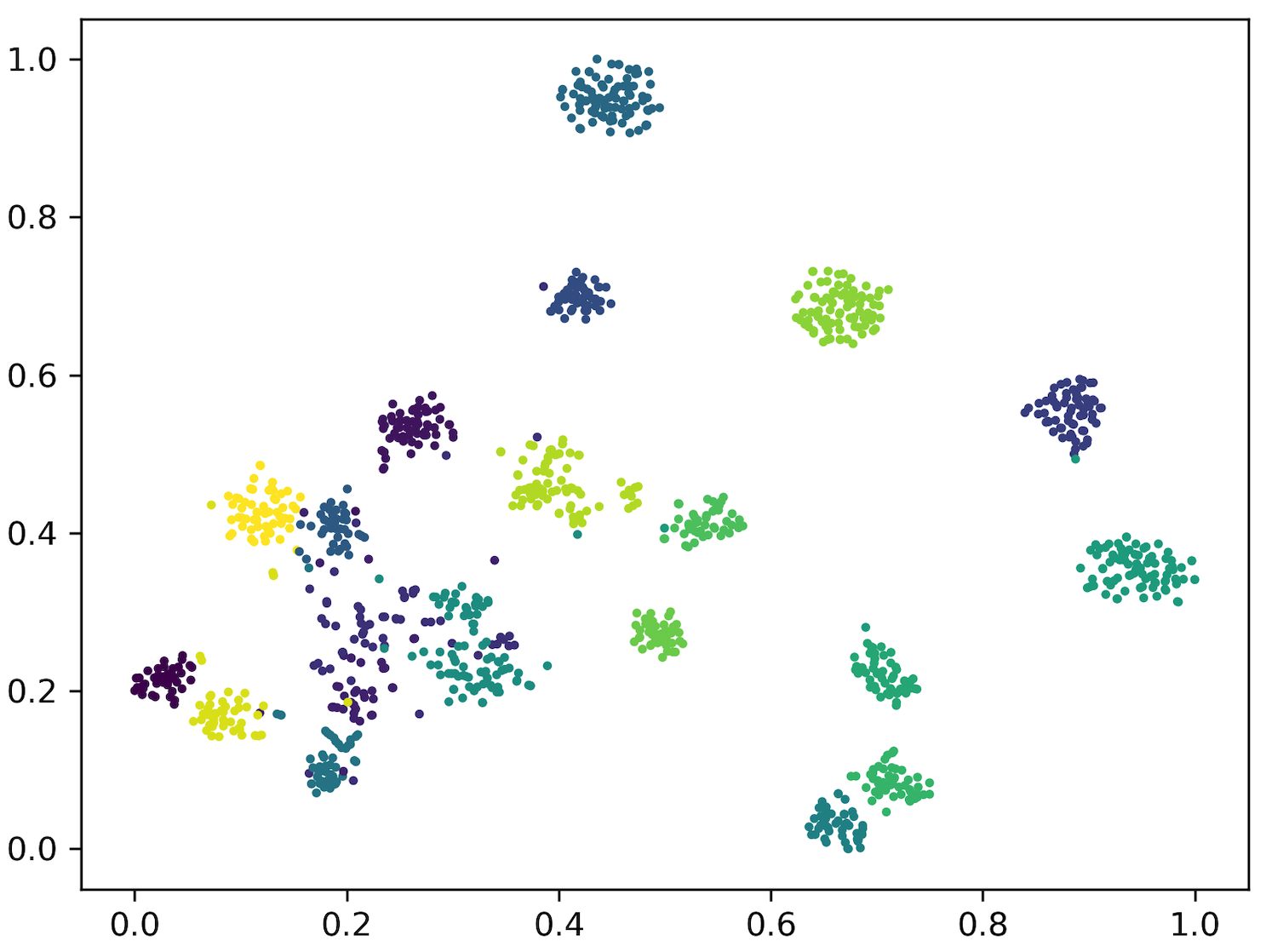}}
\subfigure[Regularized features]{
\includegraphics[width=4cm]{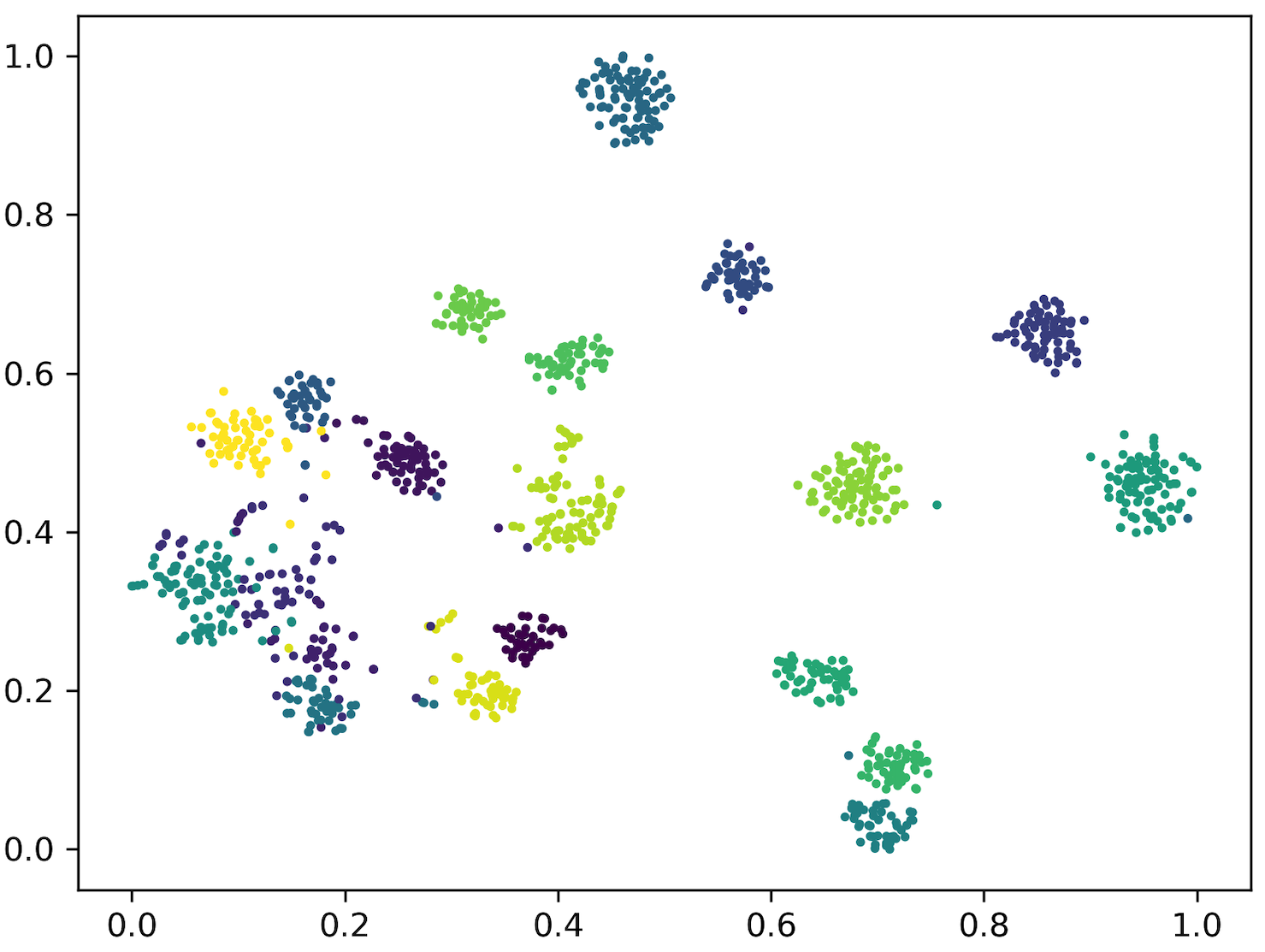}}
\caption{The visualization of four types of visual features on the unseen classes of FLO.}
\end{figure}

\begin{figure}[!htb]
\centering 
\subfigure[Naive descrip.]{
\includegraphics[width=2.7cm]{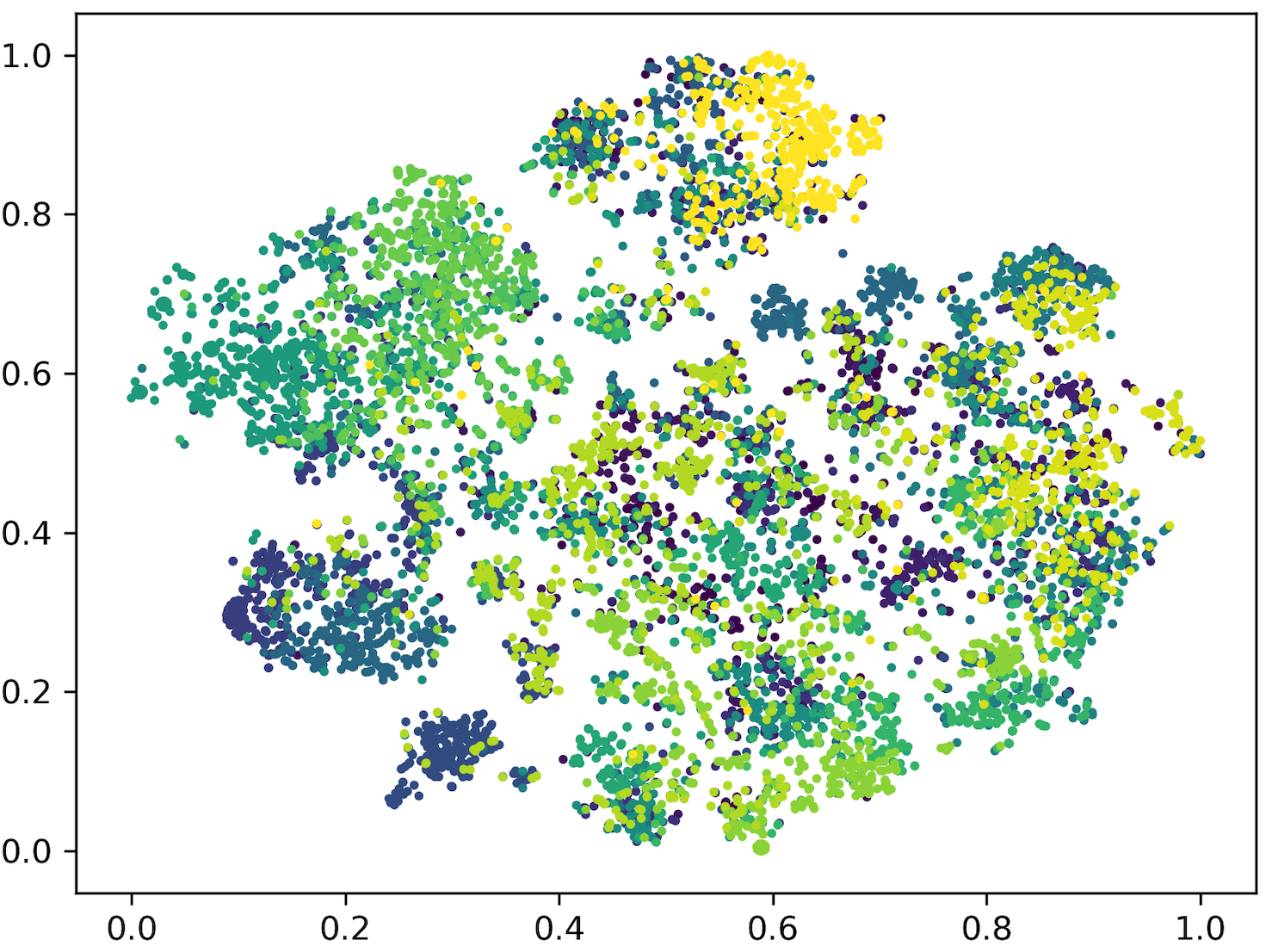}}
\subfigure[GRU descrip.]{
\includegraphics[width=2.7cm]{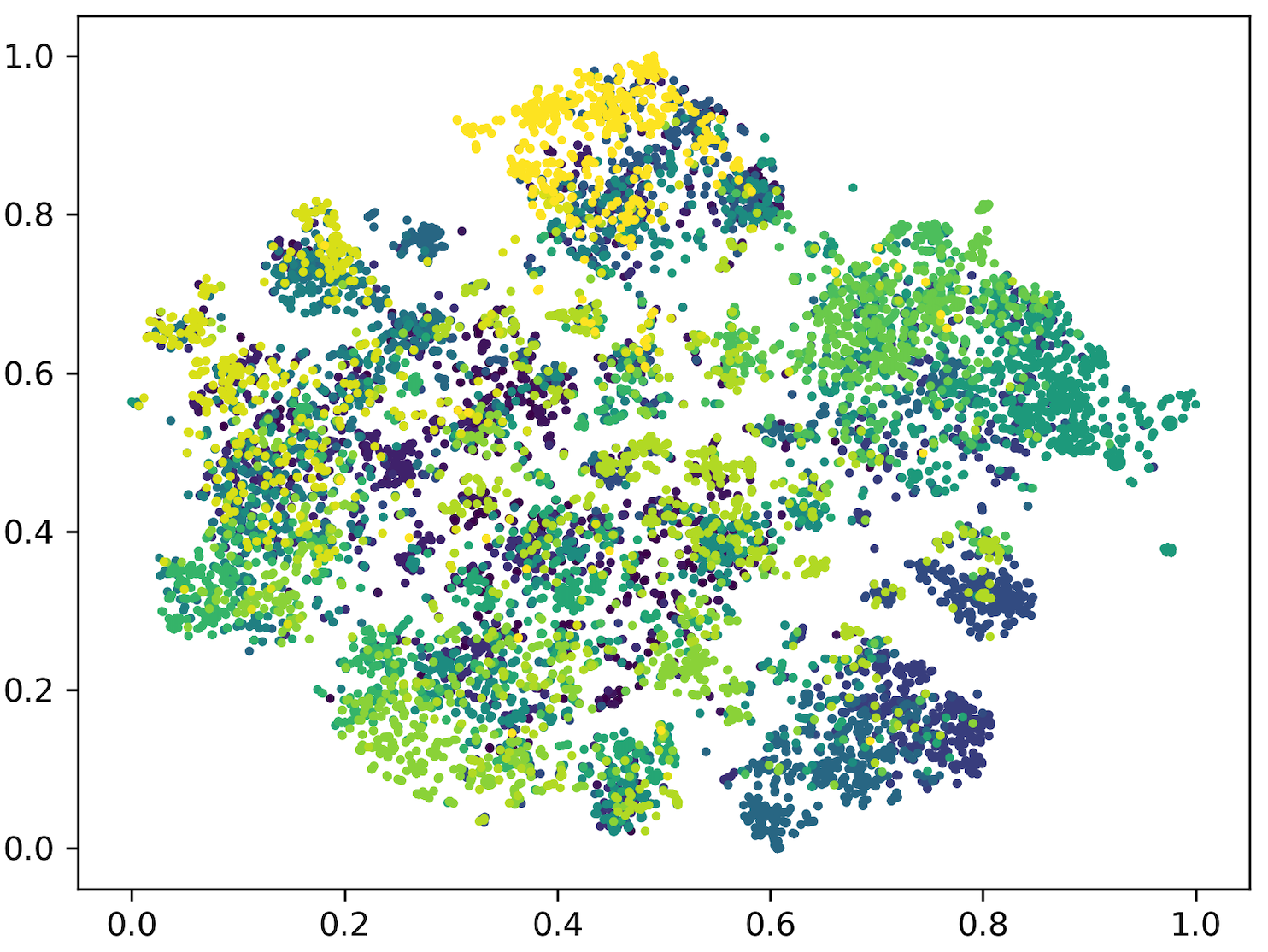}}
\subfigure[Imb. GRU descrip.]{
\includegraphics[width=2.7cm]{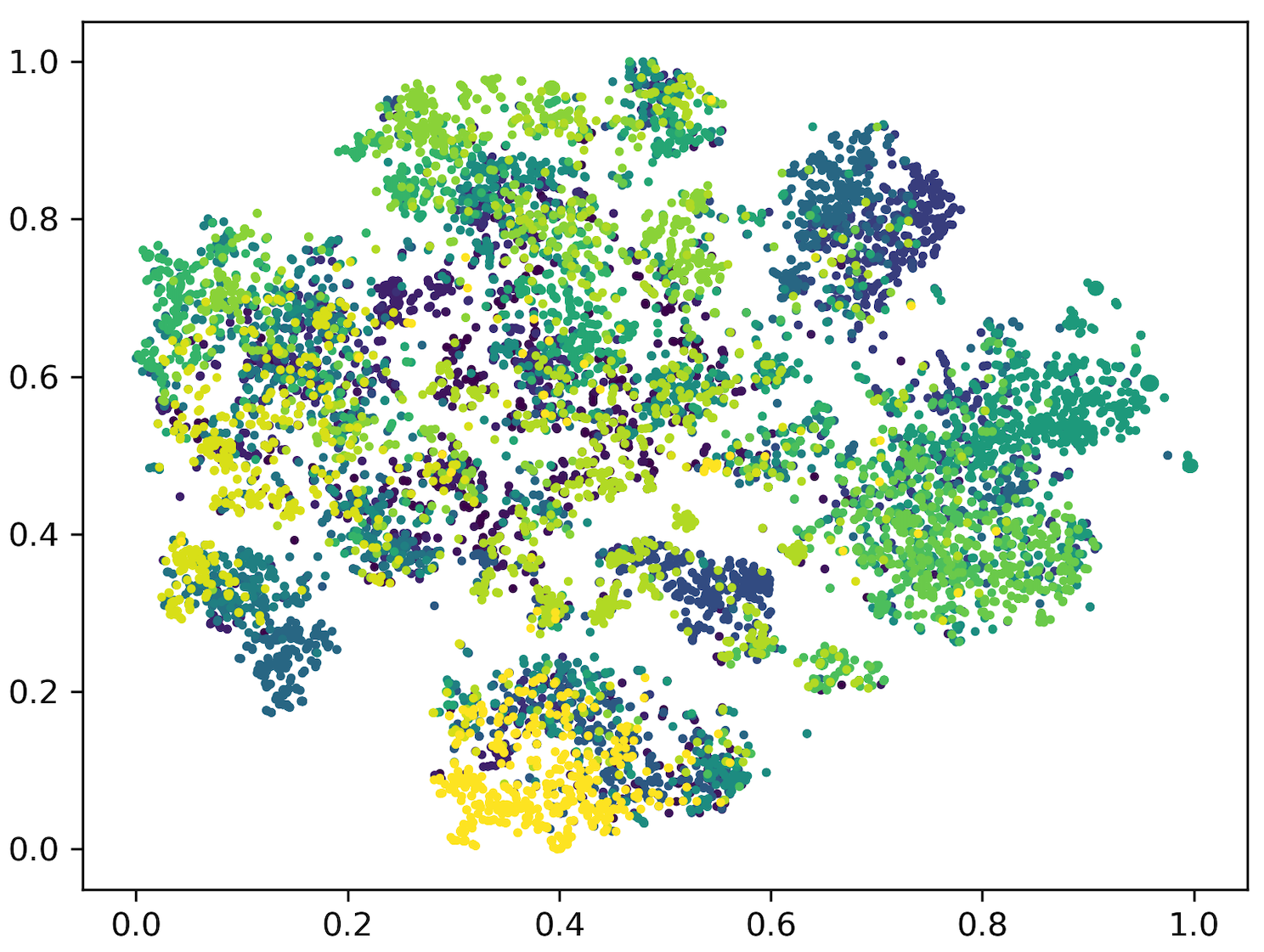}}
\caption{The sample-level visualization of three types of semantic descriptions on the unseen classes of FLO.}
\end{figure}

\subsubsection{Dataset Splits} For ZSL and GZSL, we use the proposed splits by Xian \etal \cite{27}, which strictly follow the disjoint-class principle for seen and unseen classes. Based on the splits of ZSL and GZSL, we select $N$ samples from each unseen class and add them to the training set for UFSL and GUFSL. Meanwhile, the selected unseen samples are removed from the test set. For SFSL and GSFSL, we only retain $N$ samples for each seen class in the training set. The other samples in the training set are abandoned and not added to the test set for an intuitive comparison with ZSL and GZSL. The specific splits for ZSL, GZSL, UFSL, GUFSL, SFSL, and GSFSL are listed in Table VI.
\begin{table}[!htb]\scriptsize
  \centering
      \textbf{\caption{Splits for ZSL and FSL}}
      \renewcommand\arraystretch{1.1}
      \setlength{\tabcolsep}{1.6 mm}
      \vspace{-0.7em}
      \centering
      \begin{tabular}{ccc|ccccc}
          \Xhline{1pt}
           \textbf{Task}   &  \multicolumn{2}{c}{\textbf{Scope}} &  \textbf{FLO} & \textbf{CUB}  & \textbf{SUN} & \textbf{AWA2} & \textbf{AWA}  \\
          \hline
            \multirow{4}*{\textbf{ZSL}}  & \multirow{2}*{\textbf{Tr}}   &  \#$\mathcal{X}_{s}$   &   7034   &    8821   &    12900  &    29409 &    25517 \\
                                                       &                                         &   \#$\mathcal{X}_{u}$   &    --        &    --         &       --      &    --         &    --        \\
                                                        \cline{2-8}
                                                       & \multirow{2}*{\textbf{Te}}   &  \#$\mathcal{X}_{s}$   &    --        &    --         &       --     &    --         &    --        \\
                                                       &                                         &    \#$\mathcal{X}_{u}$  &    1155   &   2967     &    1440  &    7913   &     5685  \\
          \hline
            \multirow{4}*{\textbf{GZSL}}  & \multirow{2}*{\textbf{Tr}}   &  \#$\mathcal{X}_{s}$   &    5631   &    7057 &    10320 &    23527 &    19832 \\
                                                       &                                         &   \#$\mathcal{X}_{u}$      &    --         &    --      &     --        &     --       &    --        \\
                                                      \cline{2-8}
                                                       & \multirow{2}*{\textbf{Te}}   &  \#$\mathcal{X}_{s}$   &    1403   &    1764   &    2580   &    5882 &    4958 \\
                                                       &                                         &    \#$\mathcal{X}_{u}$  &    1155   &    2967    &    1440   &    7913   &     5685 \\
          \hline
            \multirow{4}*{\textbf{UFSL}}& \multirow{2}*{\textbf{Tr}}   &  \#$\mathcal{X}_{s}$   &   7034   &    8821   &    12900  &    29409 &    25517 \\
                                                       &                                         &   \#$\mathcal{X}_{u}$   &    \begin{tiny}$Nq$\end{tiny}    &    \begin{tiny}$Nq$\end{tiny}  &   \begin{tiny}$Nq$\end{tiny}     &   \begin{tiny}$Nq$\end{tiny}  &   \begin{tiny}$Nq$\end{tiny}  \\
                                                       \cline{2-8}
                                                       & \multirow{2}*{\textbf{Te}}   &  \#$\mathcal{X}_{s}$   &    --        &    --         &       --     &    --         &    --        \\
                                                       &                                         &    \#$\mathcal{X}_{u}$  &    1155-\begin{tiny}$Nq$\end{tiny}    &   2967-\begin{tiny}$Nq$\end{tiny}     &    1440-\begin{tiny}$Nq$\end{tiny}   &    7913-\begin{tiny}$Nq$\end{tiny}   &     5685-\begin{tiny}$Nq$\end{tiny}   \\
          \hline
        \multirow{4}*{\textbf{GUFSL}}& \multirow{2}*{\textbf{Tr}}   &  \#$\mathcal{X}_{s}$  &    5631   &    7057 &    10320 &    23527 &    19832 \\
                                                       &                                         &   \#$\mathcal{X}_{u}$    &    \begin{tiny}$Nq$\end{tiny}    &    \begin{tiny}$Nq$\end{tiny}  &   \begin{tiny}$Nq$\end{tiny}     &   \begin{tiny}$Nq$\end{tiny}  &   \begin{tiny}$Nq$\end{tiny}  \\
                                                        \cline{2-8}
                                                       & \multirow{2}*{\textbf{Te}}   &  \#$\mathcal{X}_{s}$  &    1403   &    1764   &    2580   &    5882 &    4958 \\
                                                       &                                         &    \#$\mathcal{X}_{u}$  &    1155-\begin{tiny}$Nq$\end{tiny}    &   2967-\begin{tiny}$Nq$\end{tiny}     &    1440-\begin{tiny}$Nq$\end{tiny}   &    7913-\begin{tiny}$Nq$\end{tiny}   &     5685-\begin{tiny}$Nq$\end{tiny}   \\
          \hline
            \multirow{4}*{\textbf{SFSL}}& \multirow{2}*{\textbf{Tr}}   &  \#$\mathcal{X}_{s}$    &    \begin{tiny}$Np$\end{tiny}    &    \begin{tiny}$Np$\end{tiny}  &   \begin{tiny}$Np$\end{tiny}     &   \begin{tiny}$Np$\end{tiny}  &   \begin{tiny}$Np$\end{tiny}  \\
                                                       &                                         &   \#$\mathcal{X}_{u}$   &    --        &    --         &       --     &    --         &    --        \\
                                                       \cline{2-8}
                                                       & \multirow{2}*{\textbf{Te}}   &  \#$\mathcal{X}_{s}$    &    --        &    --         &       --     &    --         &    --        \\
                                                       &                                         &    \#$\mathcal{X}_{u}$  &    1155   &   2967     &    1440  &    7913   &     5685  \\
          \hline
        \multirow{4}*{\textbf{GSFSL}}& \multirow{2}*{\textbf{Tr}}   &  \#$\mathcal{X}_{s}$   &    \begin{tiny}$Np$\end{tiny}    &    \begin{tiny}$Np$\end{tiny}  &   \begin{tiny}$Np$\end{tiny}     &   \begin{tiny}$Np$\end{tiny}  &   \begin{tiny}$Np$\end{tiny}  \\
                                                       &                                         &   \#$\mathcal{X}_{u}$   &    --        &    --         &       --     &    --         &    --        \\
                                                        \cline{2-8}
                                                       & \multirow{2}*{\textbf{Te}}   &  \#$\mathcal{X}_{s}$   &    1403   &    1764   &    2580   &    5882 &    4958 \\
                                                       &                                         &    \#$\mathcal{X}_{u}$  &    1155   &    2967    &    1440   &    7913   &     5685 \\
          \Xhline{1pt}
          \end{tabular}
      \vspace{-1em}
  \end{table}

\subsubsection{Evaluation Criteria} For ZSL, UFSL, and SFSL, we evaluate the model performance with two kinds of criteria, including the average perclass top-1 accuracy, $Z$, and the training time, $ZT$. The accuracy is defined as:
\begin{equation}
\begin{aligned}
Z = \frac{1}{||\mathcal{Y}||}\sum^{||\mathcal{Y}||}_{i=1}\frac{\#\text{correct predictions in } i}{\# \text{samples in } i},
\end{aligned}
\end{equation}
and $ZT$ is defined as the time (hours) cost by a model to obtain its $Z$. For GZSL, GUFSL, and GSFSL,  we evaluate the model performance with four kinds of criteria, including the average per-class top-1 accuracy on unseen classes, $U$,  the average per-class top-1 accuracy on seen classes, $S$, the harmonic mean accuracy, $H$, and the training time, $HT$. The harmonic mean accuracy is defined as  $H = (2 * S * U) / (S+U)$, and $HT$ is defined as the time (hours) cost by a model to obtain its best $H$. In comparison with other works, we additionally use training time to evaluate the model performance, since training efficiency is as important as accuracy in practice.
\subsubsection{Models and Parameters} In this paper, we reproduce and evaluate ten EAGMs, including f-CLSWGAN, LisGAN, LsrGAN, CVAE, CADA-VAE, VAE-cFlow, f-VAEGAN-D2, tf-VAEGAN, FREE, and GCM-CF. For a fair comparison, we finetune the hyperparameters of these models for ZSL and GZSL on the five benchmark datasets. The reproduced results and the results from the original publications are compared in Table VII. As shown, a total of sixty-five results are reproduced, and the average bias is only -0.49\%. We highlight the difficulty of reproducing these networks, which are optimized by a back-propagation algorithm with numerous hyperparameters and randomness. Meanwhile, it can be observed that a total of fifteen reproduced results are better than the original results, and the average improvement is {\bf +1.24\%}, which shows that these models are reproduced well. In experiments, the hyperparameters of ZSL are applied for UFSL and SFSL without other modifications, and the hyperparameters of GZSL are applied for GUFSL and GSFSL without other modifications. Although the shared hyperparameters do not help to get the best results for  UFSL, SFSL, GUFSL, and GSFSL, they indeed contribute to presenting a fair comparison among different models and tasks.

\begin{table}[!htb]\scriptsize
  \centering
      \textbf{\caption{Comparison of Reproduced Results and Results From the Original Publications on ZSL and GZSL}}
      \renewcommand\arraystretch{1.1}
      \setlength{\tabcolsep}{0.9 mm}
      \vspace{-0.7em}
      \centering
\begin{tabular}{cc|cccccccccccccccccccccccccc}
          \Xhline{1pt}
           \multirow{2}*{\textbf{Model}}    &  \multirow{2}*{\textbf{Type}}  & \multicolumn{2}{c}{\textbf{FLO}} && \multicolumn{2}{c}{\textbf{CUB}} &&  \multicolumn{2}{c}{\textbf{SUN}} && \multicolumn{2}{c}{\textbf{AWA2}} &&  \multicolumn{2}{c}{\textbf{AWA}} \\
           \cline{3-4}   \cline{6-7}    \cline{9-10}     \cline{12-13}     \cline{15-16}   
                                                            &                                              &            Z          &           H             &&           Z          &           H             &&            Z          &           H             &&             Z          &           H              &&           Z          &           H              \\
           
            \hline
           \multirow{3}*{\textbf{f-CLS.}}   &  O    &  67.2   &    65.6   &&   57.3    &      49.7   &&     60.8     &     39.4     &&   --   &  --  &&   68.2   &   59.6  \\
                                                           &  R    &   67.6  &    68.5   &&   56.3    &      49.8   &&     60.1     &     39.2     &&   --   &  --  &&   67.4   &   59.3  \\
                                                           &   B   &   +0.4  &    +2.9   &&   -1.0    &       +0.1   &&     -0.7     &      -0.2     &&  --   &  --  &&   -0.8   &   -0.3  \\
         \hdashline[1.5pt/2pt]
           \multirow{3}*{\textbf{Lis.}}          &  O   &   69.6  &    68.3   &&   58.8    &      51.6   &&     61.7     &     40.2     &&   --  &  --  &&   70.6   &   62.3  \\
                                                              &  R    &  68.3  &    67.4   &&   57.3    &      51.1   &&     61.2     &     40.0     &&   --   & --  &&   67.9   &   61.0  \\
                                                             &   B   &   -1.3  &    -0.9    &&   -1.5     &      -0.5    &&     -0.5     &     -0.2      &&   --   &  --  &&   -2.7   &   -1.3  \\
\end{tabular}
  \end{table}

\begin{table*}[thb]\scriptsize
\centering
    \textbf{\caption{Effects of Visual features on ZSL and GZSL}}
    \renewcommand\arraystretch{1.2}
    \setlength{\tabcolsep}{0.55 mm}
    \centering
    \begin{tabular}{cc|cccccc|cccccc|cccccc|cccccc|ccccccc}    
        \Xhline{1pt}
        \multirow{2}*{\textbf{Method}}  &   \multirow{2}*{\textbf{Emb.}}  & \multicolumn{6}{c}{\textbf{FLO}} & \multicolumn{6}{c}{\textbf{CUB}} & \multicolumn{6}{c}{\textbf{SUN}} & \multicolumn{6}{c}{\textbf{AWA2}} & \multicolumn{6}{c}{\textbf{AWA}} \\
                                                         &                                                & $Z$  & $ZT$  & $U$ & $S$ & $H$  & $HT$    & $Z$  & $ZT$  & $U$ & $S$ & $H$  & $HT$   & $Z$  & $ZT$  & $U$ & $S$ & $H$  & $HT$    & $Z$  & $ZT$  & $U$ & $S$ & $H$  & $HT$     & $Z$  & $ZT$  & $U$ & $S$ & $H$  & $HT$    \\
        \hline
        \multirow{4}*{\makecell[c]{\textbf{f-CLS.}}}    &    \textbf{Orig.}    & 67.6 & \multirow{4}*{0.38}  & 61.5  & 77.2  & 68.5  & \multirow{4}*{0.34}   & 56.3 & \multirow{4}*{0.77}  & 43.9  & 57.4  & 49.8  & \multirow{4}*{1.09} & 60.1 & \multirow{4}*{1.06}  & 43.3  & 35.9  & 39.2  & \multirow{4}*{1.50} & 66.3 & \multirow{4}*{0.74}  & 47.6  & 75.2  & 58.3  & \multirow{4}*{0.87} & \multirow{4}*{67.4}  & \multirow{4}*{0.80}  & \multirow{4}*{52.7}  & \multirow{4}*{67.7}  & \multirow{4}*{59.3}   & \multirow{4}*{0.79}\\
                                                                       &    \textbf{Naiv.}    & 69.4 &        & 60.1  & 84.1  & 70.1  &         & 60.5 &       & 47.4  & 59.7  & 52.9  &       & 61.9 &        & 44.3  & 35.5  & 39.4  &         & {\color{blue}{\textbf{68.7}}} &       & {\color{red}{\textbf{64.2}}}  & 64.3  & 64.3  &       &   &      &    &   &    &     \\
                                                                       &    \textbf{Fine.}   & 71.4 &         & 67.2  & {\color{blue}{\textbf{91.8}}}  & 77.6  &         & 71.8 &       & 60.6  & {\color{blue}{\textbf{74.9}}}  & 67.0  &      & 61.3 &        & {\color{blue}{\textbf{48.2}}}  & 37.0  & 41.9  &          & 63.4 &      & 52.6  & {\color{blue}{\textbf{76.0 }}} & 62.1  &       &    &      &    &    &   &      \\
\end{tabular}
\end{table*}

\begin{table*}[!h]\scriptsize
\centering
    \textbf{\caption{Statistics of Table VIII}}
    \renewcommand\arraystretch{1.2}
    \setlength{\tabcolsep}{0.66 mm}
    \centering
    \begin{tabular}{c|cccccc|cccccc|cccccc|cccccc|ccccccc}    
        \Xhline{1pt}
        \multirow{2}*{\textbf{Items}}   & \multicolumn{6}{c}{\textbf{FLO}} & \multicolumn{6}{c}{\textbf{CUB}} & \multicolumn{6}{c}{\textbf{SUN}} & \multicolumn{6}{c}{\textbf{AWA2}} & \multicolumn{2}{c}{\textbf{AWA }}\\
                                                       & $Z$ & $ZT$  & $U$ & $S$ & $H$ & $HT$    & $Z$ & $ZT$   & $U$ & $S$ & $H$  & $HT$    & $Z$ & $ZT$   & $U$ & $S$ & $H$  & $HT$    & $Z$  & $ZT$  & $U$ & $S$ & $H$  & $HT$      & \multicolumn{2}{c}{-- }   \\
        \hline
        Best Feature     & \multicolumn{6}{c}{Reg.}  &  \multicolumn{6}{c}{Reg.}   &  \multicolumn{6}{c}{Reg.}    &  \multicolumn{6}{c}{Naiv.}    &  \multicolumn{1}{c}{Orig.}  \\
        \hline
        Original Feature's Average Accuracy   & 64.2 &  \multirow{3}*{\textbf{--}}   & 54.0  & 77.5 & 62.9  &  \multirow{3}*{\textbf{--}} & 59.2  &  \multirow{3}*{\textbf{--}}  & 46.6  & 58.4  & 51.5  & \multirow{3}*{\textbf{--}}  &  62.1  &\multirow{3}*{\textbf{--}}   & 42.7  &  36.5 &   39.2  & \multirow{3}*{\textbf{--}}   & 66.5  &   \multirow{3}*{\textbf{--}}  & 50.6  & 77.8  & 60.7 & \multirow{3}*{\textbf{--}}  &    \\
        Best Feature's Average Accuracy        & 70.4  &   & 60.3  & 93.0 & 71.6  &  & 72.7 &   & 58.9 &  74.8  & 65.0  &  & 62.1  &  & 42.7 &  38.2  & 39.7 & & 68.8  &&  52.0 &  79.5  & 61.5  &&       \multicolumn{2}{c}{\textbf{--} }   \\
               \hdashline[1.5pt/2pt]
       Accuracy Improvement                         & 6.2  &    & 6.3  & 15.5 & 8.7  & & 13.5 &   & 12.3 &  16.5  & 13.6  &  & 0.0 &   & 0.0 &  1.7  & 0.5  & & 2.3  &&   1.4 &  1.7  & 0.7  &&          \\
\end{tabular}
\end{table*}

\subsection{Embedding Evaluation on Zero-Shot Learning}
Here, we show the effects of visual and semantic features on EAGMs for ZSL and GZSL.
\subsubsection{Visual Embedding}
In last section, four types of visual features are presented, including the original features, naive features, finetuned features, and regularized features. Note that the $\alpha$, $\Delta$, and $\lambda$ in Eq. (37) and Eq. (38) for regularized features are different for each dataset. The $\alpha$ is set to 0.01 for FLO, 0.1 for CUB, 0.01 for SUN, and 0.1 for AWA2. The $\Delta$ is set to 1 for FLO, 0.01 for CUB, 1 for SUN, and 0.1 for AWA2. The $\lambda$ is set to 0.9 for FLO, 0.99 for CUB, 0.9 for SUN, and 0.9 for AWA2. Based on benchmark datasets, we perform the ZSL and GZSL tasks for EAGMs with the four types of visual features. The results are summarized in Table VIII. We analyze the results from five aspects, including the effects of visual features on a single model, the effects on different datasets, the effects on different models, the overall effects, and the model rank.\par
First, we take the results of f-CLSWGAN on FLO as an example to start our analysis. As shown in Table VIII. The change in feature embedding significantly improves both ZSL and GZSL. In comparison with the original features, for ZSL, the naive features, finetuned features, and regularized features show a $Z$ improvement of 1.8\%, 3.8\%, and 6.5\%. For GZSL, the $H$ improvement is 1.6\%, 9.1\%, and 10.2\%. {\bf Especially, the improvement presented by naive features demonstrates our effective reproduction of the original features.} And the improvement presented by finetuned features and regularized features demonstrates the contributions of adding visual and semantic information for ZSL and GZSL during feature extraction. {\bf It's shown that even simple modifications on the embedding features can improve the performance of EAGMs for ZSL remarkably.}\par
Second, we take the results of FLO, CUB, SUN, and AWA2 by f-CLSWGAN as an example to show the effects on different datasets. For ZSL, the best visual features of FLO, CUB, SUN, and AWA2 are the regularized features, regularized features, regularized features, and naive features, respectively. In comparison with the original features, the best features present a $Z$ improvement of 6.5\%, 16.5\%, 1.9\%, and 2.4\%. Meanwhile, we note that the best features of AWA2 are the naive features which demonstrates that adding visual and semantic information of seen classes during feature extraction is not always effective for the accuracy improvement of unseen classes in ZSL. For GZSL, the best visual features are regularized features, and the $H$ improvement is 10.2\%, 18.4\%, 3.2\%, and 8.7\%. We notice that the contribution of regularized features to GZSL is larger than that to ZSL, since GZSL includes seen classes at the test stage. For the training time, the $ZT$ and $HT$ are similar.\par
Third, we take the results of ten EAGMs on FLO as an example to show the effects on different models. For ZSL, the best visual features are the regularized features (f-CLS.), regularized features (Lis.), regularized features (Lsr.), finetuned features (CVAE), regularized features (CADA.), regularized features (VAE-c.), regularized features (f-VAE.), naive features (tf-VAE.), naive features (FREE), naive features (GCM). The best features present a $Z$ improvement of 6.5\%, 9.6\%, 5.5\%, 14.5\%, 6.4\%, 11.6\%, 4.4\%, 4.3\%, 0.3\%, and 7.7\%. For GZSL, the best visual features of tf-VAE. are the finetuned features, and other models' best features are the regularized features. The best features present an $H$ improvement of 10.2\%, 14.2\%, 2.7\%, 1.4\%, 9.9\%, 15.1\%, 9.8\%, 6.5\%, 3.9\%, and 14.4\%. The improvements for ZSL and GZSL demonstrate the positive effects of the given visual features on EAGMs.\par
Fourth, we list the statistics of Table VIII in Table IX for intuitive comparison. Based on the results of the ten models, we give the best visual features for each dataset, \ie, regularized features for FLO, CUB, and SUN, and naive features for AWA2. For ZSL, the best features present an average $Z$ improvement of 6.2\%, 13.5\%, 0.0\%, and 2.3\% for FLO, CUB, SUN, and AWA2. For GZSL, the best features present an average $H$ improvement of 8.8\%, 13.6\%, 0.5\%, and 0.7\%. Meanwhile, the highest accuracy for each dataset is also significantly improved by the best features. For ZSL, the best features present the highest $Z$ improvement of 4.3\%, 10.2\%, 0.0\%, and 5.2\% for FLO, CUB, SUN, and AWA2. For GZSL, the highest $H$ improvements are 6.9\%, 11.6\%, 0.1\%, and 0.3\%. These statistics show that the modifications on visual features indeed contribute to the EAGM-based ZSL and GZSL.\par

Finally, we present the ranks of different models with the best and original features, respectively, in Figure 5. When the original visual features are used, f-VAEGAN-D2 and tf-VAEGAN usually present superior performance than other models, whose average ranks are 2.1 and 2.2, respectively. When the best features are used, tf-VAEGAN and LisGAN have superior performance than other models, whose average ranks are 3.0 and 3.4, respectively. In addition, we can observe that f-CLSWGAN and LisGAN show higher ranks with the best features. But neither the best features nor the original features make CVAE have satisfactory average performance.

\begin{figure}[!htb]
\centering 
\subfigure[Original features.]{
\includegraphics[width=4.2cm]{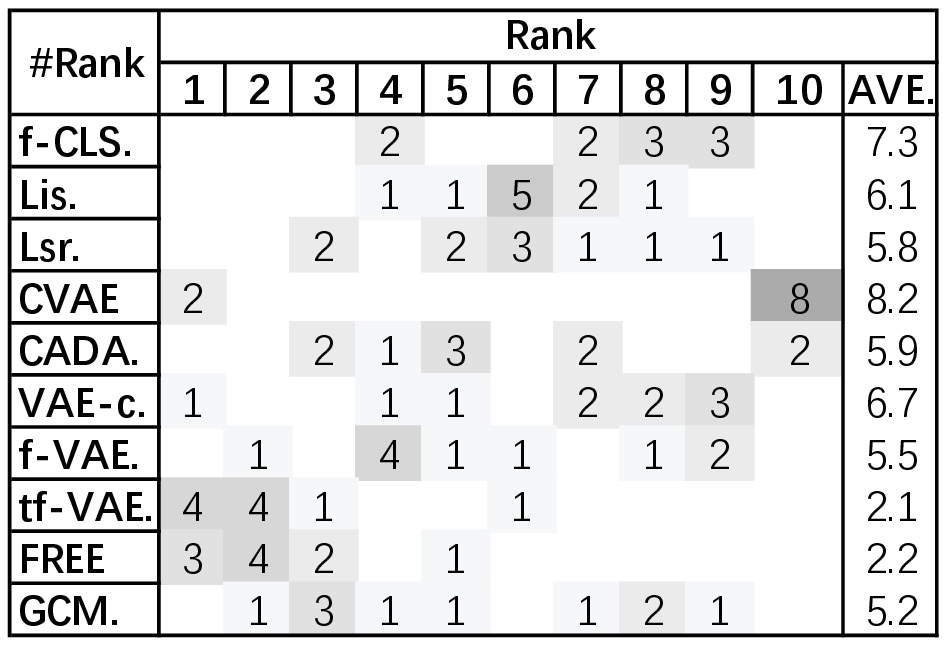}}
\subfigure[Best features]{
\includegraphics[width=4.2cm]{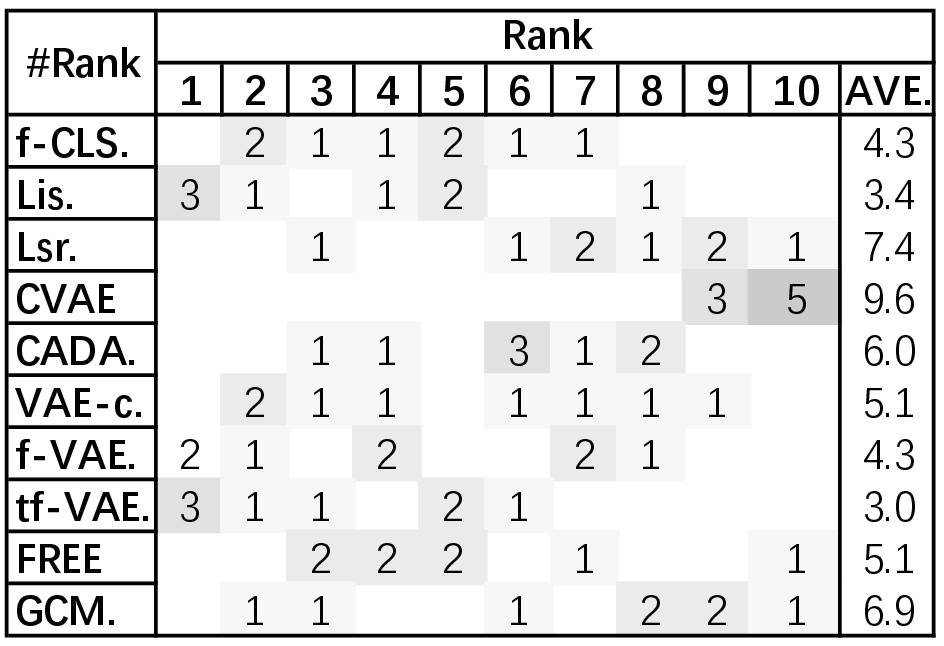}}
\caption{Ranks of ten EAGMs. For each model, the ranks of the ten results (five datasets $\times$ two tasks) from the original features are shown and the ranks of the eight results (four datasets $\times$ two tasks)  from the best features are shown.}
\end{figure}
 
\begin{table}[thb]\scriptsize
\centering
    \textbf{\caption{Effects of Semantic features on ZSL and GZSL}}
        \vspace{-1em}
    \renewcommand\arraystretch{1.1}
    \setlength{\tabcolsep}{1.25 mm}
    \centering
    \begin{tabular}{ccc|cccccc}    
        \Xhline{1pt}
        \multirow{2}*{\textbf{Type}}       & \multirow{2}*{\textbf{Method}}  &   \multirow{2}*{\textbf{Emb.}}  & \multicolumn{6}{c}{\textbf{FLO}}  \\
                                                          &                                                  &                                                & $Z$  & $ZT$  & $U$ & $S$ & $H$  & $HT$      \\
        \hline
       \multirow{12}*{\makecell[c]{\textbf{GAN}\\ \textbf{-based} \\\textbf{methods}}}    & \multirow{4}*{\makecell[c]{\textbf{f-CLSWGAN}}}    &    \textbf{Orig.}    & 74.1 & \multirow{4}*{0.13}  & 69.4  & 91.0  & 78.7  & \multirow{4}*{0.21}   \\
                                                                       &&    \textbf{Naiv.}    & 74.2 &        & 71.1  & 91.4  & 79.9  &              \\
                                                                       &&    \textbf{GRU.}   & 75.9 &         &  {\color{red}{\textbf{74.1}}}  & 89.7  & 81.2  &            \\
                                                                       &&    \textbf{Imb. GRU.}   &  {\color{blue}{\textbf{76.6}}} &         & 72.9  &  {\color{blue}{\textbf{92.0}}}  &  {\color{blue}{\textbf{81.4}}}  &            \\
       \cdashline{2-9}[1.5pt/2pt]
        &\multirow{4}*{\makecell[c]{\textbf{LisGAN}}}    &    \textbf{Orig.}    & 76.2 & \multirow{4}*{0.25}  & 71.6  & 94.8  & 81.6  & \multirow{4}*{0.34}   \\
                                                                       &&    \textbf{Naiv.}    &  79.3 &        & 71.9  & 93.8  & 81.4  &            \\
                                                                        &&    \textbf{GRU.}  & 77.6 &         & 72.9  &  {\color{blue}{\textbf{95.6}}}  & 82.8  &            \\
                                                                       &&    \textbf{Imb. GRU.}   & {\color{red}{\textbf{79.5}}} &         &  {\color{blue}{\textbf{73.7}}}  & 95.4  &  {\color{red}{\textbf{83.1}}}  &            \\
       \cdashline{2-9}[1.5pt/2pt]
        &\multirow{4}*{\makecell[c]{\textbf{LsrGAN}}}    &    \textbf{Orig.}    & 74.0 & \multirow{4}*{0.48}  & 58.5  &  {\color{blue}{\textbf{96.8}}}  & 72.9 & \multirow{4}*{0.30}   \\
                                                                       &&    \textbf{Naiv.}   & 70.9 &         & 61.7  & 94.8  & 74.7  &            \\
                                                                        &&    \textbf{GRU.}   & 72.4 &         & 63.6  & 95.4  & 76.2  &           \\
                                                                        &&    \textbf{Imb. GRU.}   &  {\color{blue}{\textbf{74.1}}} &         &  {\color{blue}{\textbf{65.3}}}  & 96.1  &  {\color{blue}{\textbf{77.7}}}  &            \\
        \hline
       \multirow{12}*{\makecell[c]{\textbf{VAE}\\ \textbf{-based} \\\textbf{methods}}}    &\multirow{4}*{\makecell[c]{\textbf{CVAE}}}     &    \textbf{Orig.}    & 66.9 & \multirow{4}*{0.05}  & 16.8  & 96.2  & 28.5 & \multirow{4}*{0.89}  \\
                                                                      & &    \textbf{Naiv.}    & 67.0 &        & 17.6  & 98.2  & 29.9  &             \\
                                                                        &&    \textbf{GRU.}  &  {\color{blue}{\textbf{70.2}}} &         & 17.7  & 98.1  & 30.0  &            \\
                                                                      &&    \textbf{Imb. GRU.}  & 70.0 &         &  {\color{blue}{\textbf{17.7}}}  &  {\color{red}{\textbf{98.3}}}  &  {\color{blue}{\textbf{30.0}}}  &            \\
       \cdashline{2-9}[1.5pt/2pt]
        &\multirow{4}*{\makecell[c]{\textbf{CADA-VAE}}}     &    \textbf{Orig.}    &  {\color{blue}{\textbf{70.4}}} & \multirow{4}*{ {\color{red}{\textbf{0.03}}}}  &  {\color{blue}{\textbf{61.9}}}  & 90.6  &  {\color{blue}{\textbf{73.5}}}  & \multirow{4}*{ {\color{red}{\textbf{0.03}}}} \\
                                                                       &&    \textbf{Naiv.}            & 67.8 &        & 51.0 &  {\color{blue}{\textbf{94.2}}}  & 66.2  &            \\
                                                                        &&    \textbf{GRU.}          & 67.6 &         & 52.9  & 94.0  & 67.7  &            \\
                                                                       &&    \textbf{Imb. GRU.}   & 70.0 &         & 53.9  & 93.9  & 68.5  &            \\
       \cdashline{2-9}[1.5pt/2pt]
        &\multirow{4}*{\makecell[c]{\textbf{VAE-cFlow}}}     &    \textbf{Orig.}    & 69.0 & \multirow{4}*{0.16}  & 64.4  & 90.5  & 75.3  & \multirow{4}*{0.17} \\
                                                                       &&    \textbf{Naiv.}          & 69.5 &        & 67.1  & 87.3  & 75.9  &             \\
                                                                        &&    \textbf{GRU.}        & 70.5 &         & 68.4  & 88.9  & 77.3  &          \\
                                                                       &&    \textbf{Imb. GRU.} &  {\color{blue}{\textbf{72.7}}} &         &  {\color{blue}{\textbf{69.1}}}  &  {\color{blue}{\textbf{91.2}}}  &  {\color{blue}{\textbf{78.6}}}  &            \\
        \hline
        
       \multirow{16}*{\makecell[c]{\textbf{VAEGAN}\\ \textbf{-based} \\\textbf{methods}}}   &\multirow{4}*{\makecell[c]{\textbf{f-VAEGAN-D2}}}    &    \textbf{Orig.}    & 69.9 & \multirow{4}*{0.24}  & 66.9  & 92.9 & 77.8 & \multirow{4}*{0.46}   \\
                                                                       &&    \textbf{Naiv.}    & 74.8 &        & 71.6  & 90.9  & 80.1  &           \\
                                                                        &&    \textbf{GRU.}  & 75.2 &         &  {\color{blue}{\textbf{73.6}}}  & 91.1  & 81.4  &            \\
                                                                       &&    \textbf{Imb. GRU.}   &  {\color{blue}{\textbf{75.9}}} &         & 72.8  &  {\color{blue}{\textbf{93.0 }}} &  {\color{blue}{\textbf{81.7}}}  &            \\
       \cdashline{2-9}[1.5pt/2pt]
        &\multirow{4}*{\makecell[c]{\textbf{tf-VAEGAN}}}    &    \textbf{Orig.}    & 70.6 & \multirow{4}*{0.23}  & 66.5  &  {\color{blue}{\textbf{92.8}}}  & 77.5  & \multirow{4}*{0.18}    \\
                                                                       &&    \textbf{Naiv.}          & 73.7 &         & 71.7  & 87.6  & 78.9  &            \\
                                                                        &&    \textbf{GRU.}        & 74.0 &         & 71.8  & 90.6  & 80.1  &            \\
                                                                       &&    \textbf{Imb. GRU.} &  {\color{blue}{\textbf{74.5}}} &         & {\color{blue}{\textbf{72.3}}}  & 92.0  &  {\color{blue}{\textbf{81.0}}}  &            \\
       \cdashline{2-9}[1.5pt/2pt]
        &\multirow{4}*{\makecell[c]{\textbf{FREE}}}    &    \textbf{Orig.}    & 70.8 & \multirow{4}*{1.83}  & 67.6  &  {\color{blue}{\textbf{93.9}}}  & 78.6  & \multirow{4}*{1.97}    \\
                                                                       &&    \textbf{Naiv.}   & 72.0 &         & 69.5  & 91.2  & 78.9  &            \\
                                                                        &&    \textbf{GRU.}  & 74.1 &         & 70.7  & 92.3  & 80.1  &             \\
                                                                       &&    \textbf{Imb. GRU.}   &  {\color{blue}{\textbf{75.6}}} &         &  {\color{blue}{\textbf{72.4}}}  & 93.4  &  {\color{blue}{\textbf{81.6}}}  &            \\
       \cdashline{2-9}[1.5pt/2pt]
        &\multirow{4}*{\makecell[c]{\textbf{GCM-CF}}}    &    \textbf{Orig.}    & 61.7 & \multirow{4}*{0.75}  & 59.9 & 90.4  & 72.1  & \multirow{4}*{0.72}   \\
                                                                       &&    \textbf{Naiv.}    & 67.5 &         & 65.1  &  {\color{blue}{\textbf{91.0}}}  & 75.9  &            \\
                                                                        &&    \textbf{GRU.} &  {\color{blue}{\textbf{69.0}}} &         & 65.2  & 90.5  & 75.8  &            \\
                                                                       &&    \textbf{Imb. GRU.} & 68.3 &         &  {\color{blue}{\textbf{65.8}}}  & 89.8  &  {\color{blue}{\textbf{76.0}}}  &            \\
        \Xhline{1pt}
        \end{tabular}
\end{table}

\subsubsection{Semantic Embedding} For FLO, four types of semantic features are provided, including the original descriptions, naive descriptions, GRU-based descriptions, and imbalanced GRU-based descriptions. Based on the regularized visual features, we explore the effects of semantic features on the performance of EAGMs for ZSL and GZSL. Note that the $\alpha$ in Eq. (42) is set to 0.7 for FLO. The performance evaluation is summarized in Table X. Similarly, we take the results of f-CLSWGAN as an example to start our analysis. The change in semantic features significantly improves both ZSL and GZSL. For ZSL, the naive descriptions, GRU-based descriptions, and imbalanced GRU-based descriptions show a $Z$ improvement of 0.1\%, 1.8\%, and 2.5\% in comparison with the original descriptions. For GZSL, the $H$ improvements are 1.2\%, 2.5\%, and 2.7\%. Meanwhile, six EAGMs obtain their best performance for ZSL and nine EAGMs obtain their best performance for GZSL with the designed imbalanced GRU-based features. In comparison with the original descriptions, the imbalanced GRU-based descriptions averagely improve $Z$ for ZSL by 3.3\% and improve $H$ for GZSL by 2.3\% on the ten EAGMs. {\bf These results demonstrate that our modifications on semantic descriptions indeed contribute to EAGM-based ZSL and GZSL.}\par
\subsection{Baseline of Few-Shot Learning}
Here, we show a strong semantically guided baseline for EAGM-based FSL, including UFSL, GUFSL, SFSL, and GSFSL. The paradigm shown in Eqs. (4)-(6) is used to apply EAGMs for FSL. Five benchmark datasets are used, including FLO, CUB, SUN, AWA2, and AWA. The features used for FSL are listed in Table XI. As shown, for UFSL and GUFSL, we use the best visual features defined in Table IX for each dataset, which are helpful to obtain satisfactory performance. For SFSL and GSFSL, we use the original or naive visual features to follow the few seen sample principle in the training stage. In addition, we uniformly use the original semantic features for the four few-shot tasks. The few-shot scenarios where one sample, five samples, ten samples, and twenty samples are used for each class are considered. The results of UFSL, GUFSL, SFSL, and GSFSL are summarized in Table XII. We analyze the results from five aspects, including the performance of a single model, the performance on different datasets, the visualization comparison, the model rank, and the effects of imbalanced GRU-based descriptions.\par

\begin{table}[htb]\scriptsize
  \centering
      \textbf{\caption{features Used for the Five Benchmark Datasets in FSL}}
      \renewcommand\arraystretch{1.2}
      \setlength{\tabcolsep}{2.9 mm}
      \vspace{-0.7em}
      \centering
\begin{tabular}{cc|cccccccccccccccccccccccccc}
          \Xhline{1pt}
           \textbf{Task}    &  \textbf{Emb.}  & \textbf{FLO} &  \textbf{CUB} & \textbf{SUN} & \textbf{AWA2} &  \textbf{AWA} \\
            \hline
           \multirow{2}*{\textbf{UFSL/GUFSL}}   &  $\mathcal{X}$     &  Reg.   &    Reg.   &  Reg.    &      Naiv. &   Orig.     \\
                                                                       &  $\mathcal{A}$     &  Orig.  &    Orig.  &  Orig.    &     Orig.    &   Orig.     \\
         \hdashline[1.5pt/2pt]
           \multirow{2}*{\textbf{SFSL/GSFSL}}    & $\mathcal{X}$      &  Naiv.  &   Naiv.   &   Orig.    &     Naiv.   &   Orig.     \\
                                                                        &  $\mathcal{A}$    &  Orig.  &    Orig.  &  Orig.    &     Orig.    &   Orig.     \\                   
          \Xhline{1pt}
          \end{tabular}
  \vspace{-1em}
  \end{table}

  \begin{table*}[!th]\tiny
    \centering
        \textbf{\caption{Results of UFSL, GUFSL, SFSL, and GSFSL}}
            \vspace{-1em}
        \renewcommand\arraystretch{1.2}
        \setlength{\tabcolsep}{1 mm}
        \centering
        \begin{tabular}{ccc|cccccc|cccccc|cccccc|cccccc|ccccccc}    
            \Xhline{1pt}
            \multirow{2}*{\textbf{Method}}&   \multirow{2}*{\textbf{Task}}   &   \multirow{2}*{\textbf{Shots}}  & \multicolumn{6}{c}{\textbf{FLO}} & \multicolumn{6}{c}{\textbf{CUB}} & \multicolumn{6}{c}{\textbf{SUN}} & \multicolumn{6}{c}{\textbf{AWA2}} & \multicolumn{6}{c}{\textbf{AWA}} \\
                                                           &  &                                                & $Z$  & $ZT$  & $U$ & $S$ & $H$  & $HT$    & $Z$  & $ZT$  & $U$ & $S$ & $H$  & $HT$   & $Z$  & $ZT$  & $U$ & $S$ & $H$  & $HT$    & $Z$  & $ZT$  & $U$ & $S$ & $H$  & $HT$     & $Z$  & $ZT$  & $U$ & $S$ & $H$  & $HT$    \\
            \hline
            \multirow{8}*{\makecell[c]{\textbf{f-CLS.}}}& \multirow{4}*{\makecell[c]{\textbf{U}}}    &    \textbf{1}    & 75.0 & \multirow{4}*{0.68}  & 69.9  & 91.6  & 79.3  & \multirow{4}*{{\color{red}{\textbf{0.04}}}}   & 72.6 & \multirow{4}*{2.04}  & 63.3 & 74.1  & 68.3  & \multirow{4}*{1.88} & 62.1 & \multirow{4}*{2.19}  & 46.6 & 39.0  & 42.5 & \multirow{4}*{2.14} & 67.7 & \multirow{4}*{1.09}  & 58.5  & 75.6  & 66.0  & \multirow{4}*{0.73} & 65.5 & \multirow{4}*{1.57} & 52.3  & 67.0  & 58.7   & \multirow{4}*{1.63}\\
                                                                  &         &    \textbf{5}    & 75.3 &        & 81.5   & 97.1   & 88.6   &         & 74.5  &       & 64.4   & {\color{red}{\textbf{80.9}}}   & 71.7   &       & 63.3  &        & 49.3   & 39.2 & 44.0  &         & 68.0 &       & 58.2   & 75.9 & 65.9   &       & 64.8   &    & 51.4  & 69.5  & 59.1   &  \\
                                                                   &        &    \textbf{10}  & 89.7 &        & 91.8   & 96.6   & 94.1   &         & 77.7  &       & 71.2   & 79.9   & 75.3   &       & 66.1  &        & 55.4   & 39.4  & 46.0  &         & 72.6 &       & 55.4   & 77.4 & 64.6   &       & 65.1   &    & 52.5  & 67.7  & 59.2   &  \\
\end{tabular}
    \end{table*}

First, we take the results of f-CLSWGAN on FLO as an example to start our analysis. In comparison with the zero-shot scenarios, UFSL and GUFSL improve model performance with a few unseen samples. Specifically, for UFSL, the one-shot, five-shot, ten-shot, and twenty-shot scenarios present a $Z$ improvement of 0.9\%, 1.2\%, 15.6\%, and 21.4\%. For GUFSL, the $H$ improvement is 0.6\%, 9.9\%, 15.4\%, and 17.4\%. {\bf Usually, the accuracy is improved with the increasing number of unseen samples}. The accuracy improvement by more training samples can also be observed on SFSL and GSFSL. In comparison with the one-sample per-seen class scenario, the five-shot, ten-shot, and twenty-shot scenarios present a $Z$ improvement of 12.1\%, 19.7\%, and 18.0\% for SFSL. The $H$ improvement for GSFSL is 36.1\%, 48.2\%, and 51.5\%. {\bf However, the performance of SFSL and GSFSL is inferior to the performance of UFSL, GUFSL, ZSL, and GZSL, since SFSL and GSFSL have only a few seen samples for model training, which are more challenging}.\par

Second, we observe the performance of f-CLSWGAN on FLO, CUB, SUN, AWA2, and AWA. Similarly, the accuracies of UFSL, GUFSL, SFSL, and GSFSL on the five benchmark datasets are usually improved with the increasing number of samples. From the one-shot learning to the twenty-shot or ten-shot (SUN) learning, the $Z$ improvement of UFSL on FLO, CUB, SUN, AWA2, and AWA is 20.5\%, 9.6\%, 4.0\%, 2.4\%, and 0.1\%, and the $H$ improvement for GUFSL is 16.8\%, 10.9\%, 3.5\%, 1.1\%, and 2.3\%. However, the improvements on SUN, AWA2 and AWA are not as significant as the improvements on FLO and CUB. For AWA2 and AWA, more significant improvements can be observed on other models, like LsrGAN and CADA-VAE. For SUN, the little improvements may be caused by the numerous classes, which make the recognition task more challenging. In addition, the accuracy improvements of SFSL and GSFSL by increasing the number of samples are significant on all the datasets, since the one-shot per-seen sample scenario produces a weak baseline. \par

\begin{figure*}[!htb]
\centering 
\includegraphics[width=0.6\textwidth]{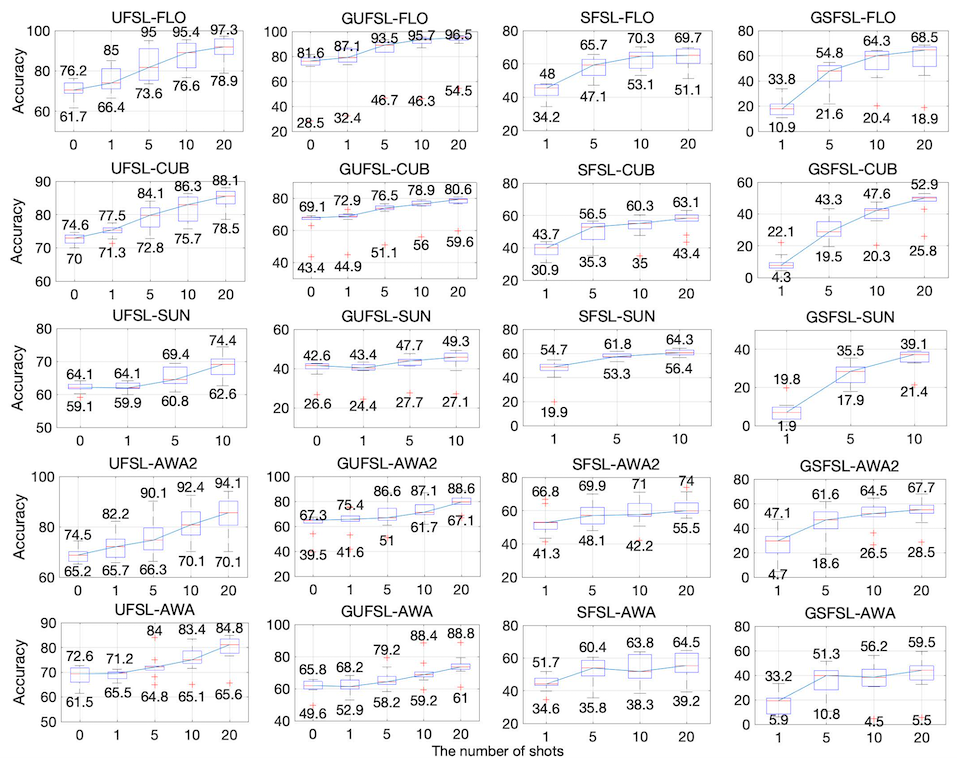} 
\caption{Visualization of the performance of ten EAGMs for UFSL, GUFSL, SFSL, and GSFSL on the five benchmark datasets.}
\end{figure*}

Third, we visualize the performance of ten EAGMs for UFSL, GUFSL, SFSL, and GSFSL on the five benchmark datasets in Figure 6. As shown, the performance of EAGMs deteriorates in order for the four tasks. {\bf In other words, the difficulty of the four tasks increases in turn}. For example, the best performance of UFSL and GUFSL is about 85\% $\sim$ 98\% and 80\% $\sim$ 97\%, respectively. And the best performance of SFSL and GSFSL is about 60\% $\sim$ 75\% and 40\% $\sim$ 70\%, respectively. More models and techniques can be designed to address SFSL and GSFSL. {\bf The results shown in Figure 6 also demonstrate that increasing the number of samples in FSL could generally improve the performance of EAGMs}. In addition, we can find that there are some outliers in the visualization of the results of GUFSL in
Figure 6, which shows degraded performance in comparison with others. These outliers are caused by CVAE, which is a naive generator and does not take classification ability and visual-semantic matching into consideration.\par

Fourth, we present the ranks of different models for UFSL, GUFSL, SFSL, and GSFSL in Figure 7. For UFSL, CADA-VAE, LsrGAN, and VAE-cFlow usually present superior performance than other models, whose average ranks are 1.8, 3.4, and 3.4. For GUFSL, CADA-VAE and LsrGAN also obtain satisfactory performance, whose average ranks are 1.6 and 2.9, respectively. For SFSL, the best models are VAE-cFlow and tf-VAEGAN, whose average ranks are  4.0 and 3.5, respectively. For GSFSL, the best models are CADA-VAE and tf-VAEGAN, whose average ranks are 2.1 and 3.5, respectively. As listed in Table III, all of LsrGAN, CADA-VAE, tf-VAEGAN have a regularization on visual-semantic matching, which may contribute to the performance of few-shot learning.\par

\begin{figure}[!htb]
\centering 
\subfigure[UFSL]{
\includegraphics[width=4.2cm]{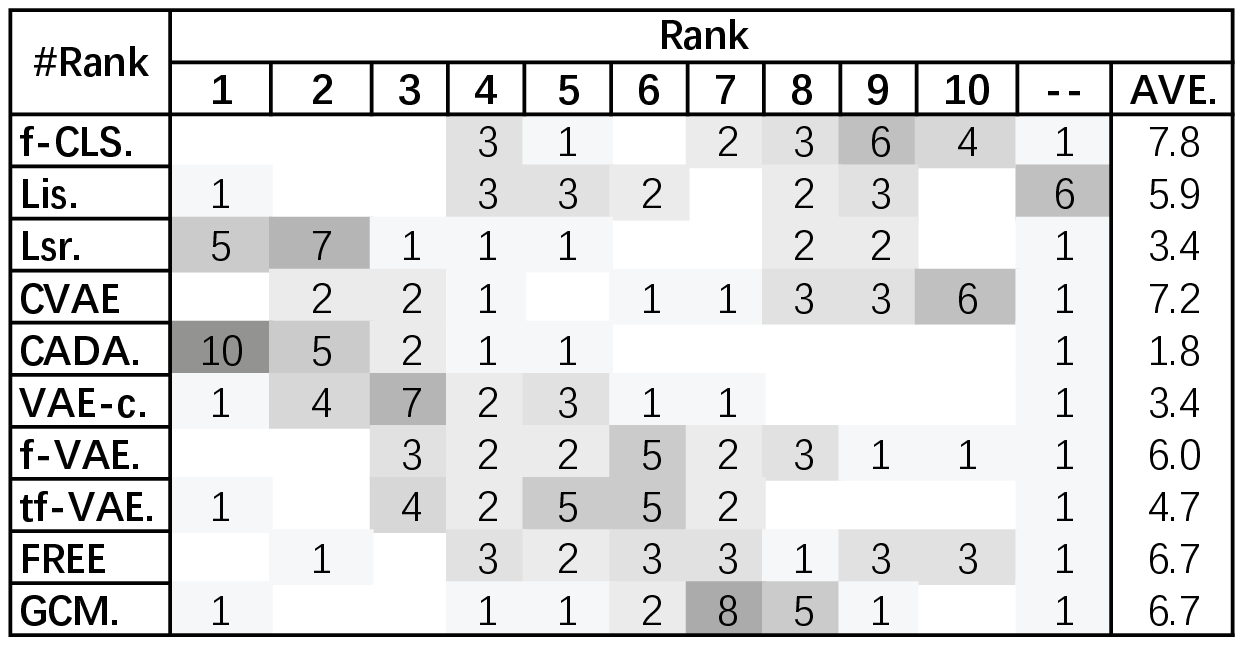}}
\subfigure[GUFSL]{
\includegraphics[width=4.2cm]{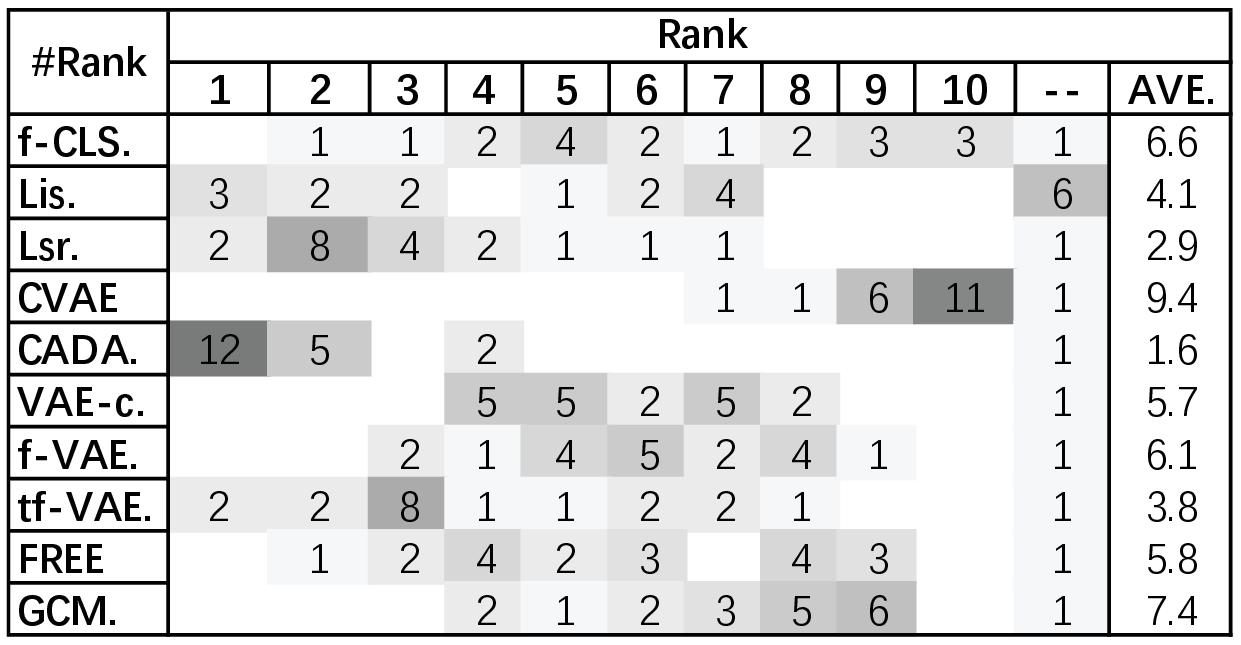}}
\subfigure[SFSL]{
\includegraphics[width=4.2cm]{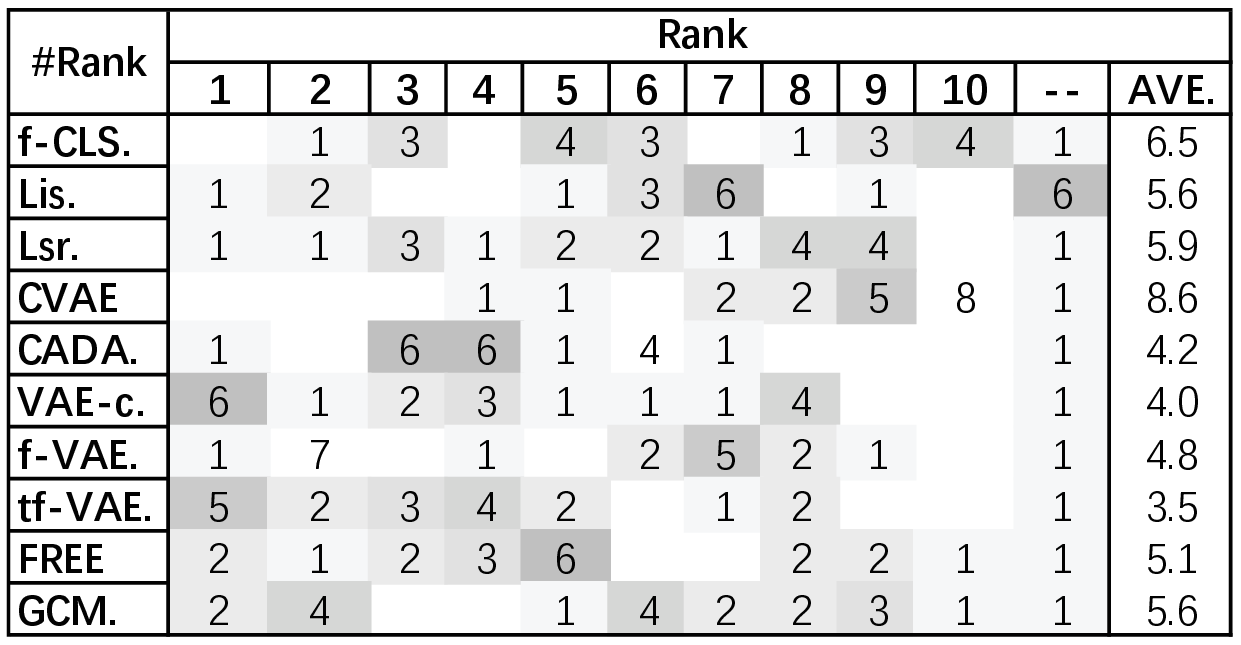}}
\subfigure[GSFSL]{
\includegraphics[width=4.2cm]{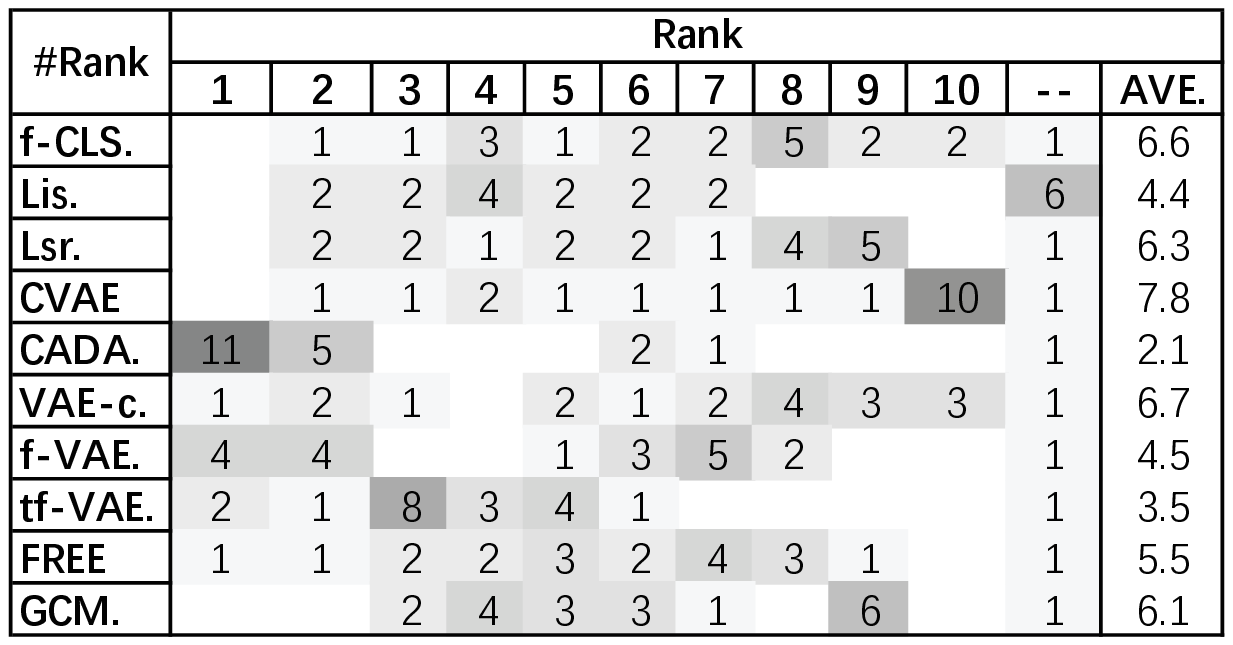}}
\caption{Ranks of ten EAGMs for FSL. For each model, the ranks of the twenty results (five datasets $\times$ four scenarios of different shots) of each task are shown, and ``- -" denotes the result is missing.}
\vspace{-1em}
\end{figure}

Finally, we use the designed imbalanced GRU-based descriptions of FLO for few-shot learning. The results are summarized in Table XIII. {\bf Through comparing the results in Table XIII and the results in Table XII, we can find that the designed imbalanced GRU-based descriptions indeed improve the performance of EAGMS for few-shot learning}. Generally,  an accuracy improvement of  3\% $\sim$ 6\% for few-shot learning and generalized few-shot learning can be observed. So the improvement of semantic features deserves demanding attention. 
\begin{table}[!th]\tiny
\centering
    \textbf{\caption{Results of FSL with Imbalanced GRU-based Descriptions on FLO}}
     \vspace{-1em}
    \renewcommand\arraystretch{1.2}
    \setlength{\tabcolsep}{1.9 mm}
    \centering
    \begin{tabular}{cccc|cccccc}    
        \Xhline{1pt}
       \multirow{2}*{\textbf{Type}}      & \multirow{2}*{\textbf{Method}}&   \multirow{2}*{\textbf{Task}}   &   \multirow{2}*{\textbf{Shots}}  & \multicolumn{6}{c}{\textbf{FLO}}  \\
                                                       &  &  &                                                & $Z$  & $ZT$  & $U$ & $S$ & $H$  & $HT$     \\
        \hline
         \multirow{24}*{\makecell[c]{\textbf{GAN}\\ \textbf{-based} \\\textbf{methods}}}    &\multirow{8}*{\makecell[c]{\textbf{f-CLSWGAN}}}& \multirow{4}*{\makecell[c]{\textbf{(G)UFSL}}}    &    \textbf{1}    & 77.2 & \multirow{4}*{1.08}  & 72.8  & 93.8  & 82.0  & \multirow{4}*{1.37} \\
                                                              &&         &    \textbf{5}    & 87.1 &        & 87.0   & 96.5   & 91.5   &   \\
                                                               &&        &    \textbf{10}  & 95.0 &        & 94.4   & 97.4   & 95.9   &  \\
                                                                &&        &    \textbf{20} & 95.8 &        & 95.8   & 97.4   & 96.6  &  \\
       \cdashline{3-10}[1.5pt/2pt]       
       && \multirow{4}*{\makecell[c]{\textbf{(G)SFSL}}}     &    \textbf{1}     & {\color{blue}{\textbf{50.9}}} & \multirow{4}*{0.19}  & 36.5  & 20.9 & 13.4  & \multirow{4}*{26.5}\\
                                                               &&                     &    \textbf{5}     & 65.8 &        & 53.0   & 53.4  & 53.2     &       \\
                                                                &&                   &    \textbf{10}    & 66.9 &        & 61.6   & 62.9   & 62.2   &     \\
                                                                 &&                  &    \textbf{20}   & 68.7 &        & 61.2   & 74.1   & 67.0   &    \\
        \cline{2-10}
        &\multirow{8}*{\makecell[c]{\textbf{LisGAN}}}& \multirow{4}*{\makecell[c]{\textbf{(G)UFSL}}}    &    \textbf{1}    & -- & \multirow{4}*{1.45}  & --  & --  & --  & \multirow{4}*{1.66}\\
                                                                 &&         &    \textbf{5}     & 86.4 &        & 86.0   &  97.8  & 91.5   &  \\
                                                                 &&        &    \textbf{10}    & 89.5 &        & 91.5   & 97.7   & 94.5  &  \\
                                                                 &&        &    \textbf{20}    & 92.5 &        & 94.3   & 97.2   & 95.7  &   \\
       \cdashline{3-10}[1.5pt/2pt]        
      & & \multirow{4}*{\makecell[c]{\textbf{(G)SFSL}}}                  &    \textbf{1}     & -- & \multirow{4}*{1.43}  & --  & --  & --  & \multirow{4}*{1.29}\\
                                                                 &&        &    \textbf{5}     & {\color{blue}{\textbf{65.9}}} &       & 59.0  & 52.3  & 55.5  &   \\
                                                                 &&       &    \textbf{10}    &   {\color{blue}{\textbf{69.0}}} &       & 61.8   &  {\color{blue}{\textbf{68.3}}}   &  {\color{blue}{\textbf{64.9}}} &    \\
                                                                 &&       &    \textbf{20}   & 68.2 &        & 60.4   & 76.7   & 67.6  &  \\
        \cline{2-10}
        &\multirow{8}*{\makecell[c]{\textbf{LsrGAN}}}& \multirow{4}*{\makecell[c]{\textbf{(G)UFSL}}}    &    \textbf{1}  &84.8 & \multirow{4}*{1.66}  &  {\color{red}{\textbf{82.0}}}  & 94.8  & {\color{red}{\textbf{88.0}}} & \multirow{4}*{1.73} \\
                                                              &&         &    \textbf{5}    & {\color{red}{\textbf{92.6}}} &           & 89.3   & 96.4   & 92.7   &  \\
                                                               &&        &    \textbf{10}    & 95.6 &         & 93.3   &  {\color{red}{\textbf{98.2}}}  & 95.7  &   \\
                                                                &&        &    \textbf{20}    & {\color{red}{\textbf{98.2}}} &        & {\color{red}{\textbf{96.6}}}   & 97.8   & {\color{red}{\textbf{97.2}}}   &  \\
       \cdashline{3-10}[1.5pt/2pt]           
      && \multirow{4}*{\makecell[c]{\textbf{(G)SFSL}}}    &    \textbf{1}     & 47.4 & \multirow{4}*{0.24}  & 34.0  & 28.5  & 31.0  & \multirow{4}*{0.74}  \\
                                                               &&        &    \textbf{5}     & 63.7 &        & 52.5   & 60.7   &  {\color{blue}{\textbf{56.3}}}    & \\
                                                                &&       &    \textbf{10}    & 67.6 &        &  {\color{blue}{\textbf{64.0}}}   & 63.4   & 63.7  & \\
                                                                 &&       &    \textbf{20}   & 69.0 &        & 63.3   & 72.4   & 67.5    & \\
        \hline
         \multirow{24}*{\makecell[c]{\textbf{VAE}\\ \textbf{-based} \\\textbf{methods}}}    &\multirow{8}*{\makecell[c]{\textbf{CVAE}}}& \multirow{4}*{\makecell[c]{\textbf{(G)UFSL}}}    &    \textbf{1}    & 74.5 & \multirow{4}*{0.05}  & 32.1  & {\color{red}{\textbf{98.0}}} & 48.4  & \multirow{4}*{0.78} \\
                                                              &&         &    \textbf{5}       & 84.6 &        & 51.4   & {\color{red}{\textbf{98.2}}}   & 67.5   &  \\
                                                               &&        &    \textbf{10}     & 88.3 &        & 64.2   & 98.0   & 77.6   &  \\
                                                                &&        &    \textbf{20}    & 90.5 &        & 73.2   & {\color{red}{\textbf{97.9}}}   & 83.7   & \\
       \cdashline{3-10}[1.5pt/2pt]          
       && \multirow{4}*{\makecell[c]{\textbf{(G)SFSL}}}                  &    \textbf{1}     & 35.0 & \multirow{4}*{0.09}  & 14.2  &  {\color{blue}{\textbf{43.1}}}  & 21.4 & \multirow{4}*{2.21}  \\
                                                              & &        &    \textbf{5}     & 54.4 &        & 16.3   & 64.1   & 26.0   &  \\
                                                                &&       &    \textbf{10}    & 53.1 &        & 17.4   & 67.3   & 27.7  & \\
                                                                 &&       &    \textbf{20}   & 54.1 &        & 13.7   & 70.2   & 23.0   &  \\
        \cline{2-10}
        &\multirow{8}*{\makecell[c]{\textbf{CADA-VAE}}}& \multirow{4}*{\makecell[c]{\textbf{(G)UFSL}}}    &    \textbf{1}    &  {\color{red}{\textbf{85.9}}} & \multirow{4}*{ {\color{red}{\textbf{0.05}}}}  & 76.7  &  96.8 &  85.6 & \multirow{4}*{ {\color{red}{\textbf{0.06}}}}  \\
                                                              &&         &    \textbf{5}    & 91.4 &        & 89.6   & 96.9   & 93.1 &  \\
                                                               &&        &    \textbf{10}    & 94.2 &        & 92.0   & 97.1   & 94.5   &  \\
                                                                &&        &    \textbf{20}    & 95.1 &        & 94.1   & 96.4   & 95.2  & \\
       \cdashline{3-10}[1.5pt/2pt]           
       && \multirow{4}*{\makecell[c]{\textbf{(G)SFSL}}}                  &    \textbf{1}     & 40.5 & \multirow{4}*{ {\color{blue}{\textbf{0.01}}}}  & 33.5  & 38.6 &  {\color{blue}{\textbf{35.9}}} & \multirow{4}*{ {\color{blue}{\textbf{0.01}}}} \\
                                                               &&        &    \textbf{5}     & 58.3 &        & 46.7   &  {\color{blue}{\textbf{66.2}}}   & 54.7  &  \\
                                                                &&       &    \textbf{10}    & 61.8 &        & 53.2   & 65.4   & 58.7  &  \\
                                                                 &&       &    \textbf{20}   & 62.0 &        & 47.5   &  {\color{blue}{\textbf{77.5}}}   & 58.9 &  \\
        \cline{2-10}
       & \multirow{8}*{\makecell[c]{\textbf{VAE-cFlow}}}& \multirow{4}*{\makecell[c]{\textbf{(G)UFSL}}}    &    \textbf{1}    & 77.6 & \multirow{4}*{0.47}  & 69.7  & 93.2  & 79.7 & \multirow{4}*{0.47} \\
                                                              &&         &    \textbf{5}    & 89.5 &        & 85.6   & 94.4   & 89.8   & \\
                                                               &&        &    \textbf{10}    & 92.4 &        & 90.1   & 94.7   & 92.3 &   \\
                                                                &&        &    \textbf{20}    & 94.2 &        & 94.1   & 94.2   & 94.1  &   \\
       \cdashline{3-10}[1.5pt/2pt]          
       && \multirow{4}*{\makecell[c]{\textbf{(G)SFSL}}}                  &    \textbf{1}     & 45.6 & \multirow{4}*{0.45}  & 31.5  & 5.5  & 9.4 & \multirow{4}*{0.45}  \\
                                                               &&        &    \textbf{5}     & 57.8 &        & 56.5   & 28.2   & 37.6  &   \\
                                                                &&       &    \textbf{10}    & 60.2 &        & 54.5   & 44.8   & 49.1  &   \\
                                                                 &&       &    \textbf{20}   & 62.5 &        & 55.2   & 51.7   & 53.4  &   \\
        \hline
       \multirow{32}*{\makecell[c]{\textbf{VAEGAN}\\ \textbf{-based} \\\textbf{methods}}}    & \multirow{8}*{\makecell[c]{\textbf{f-VAEGAN-D2}}}& \multirow{4}*{\makecell[c]{\textbf{(G)UFSL}}}    &    \textbf{1}    & 76.7 & \multirow{4}*{2.85}  & 75.2 & 92.5  & 83.0 & \multirow{4}*{3.89}  \\
                                                              &&          &    \textbf{5}    & 78.1 &        & 76.5   & 95.8   & 85.1 &          \\
                                                               &&         &    \textbf{10}    & 85.8 &        & 86.3   & 96.4   & 91.0  &         \\
                                                                &&        &    \textbf{20}    & 92.7 &        & 93.1   & 96.8   & 94.9  &         \\
       \cdashline{3-10}[1.5pt/2pt]      
       && \multirow{4}*{\makecell[c]{\textbf{(G)SFSL}}}                  &    \textbf{1}   & 50.0 & \multirow{4}*{1.80}  & 22.2  & 11.9  & 15.5  & \multirow{4}*{1.39}  \\
                                                               &&        &    \textbf{5}     & 62.5 &         & 58.0   & 38.7   & 46.4   &           \\
                                                                &&       &    \textbf{10}    & 67.6 &        & 58.6   & 66.7   & 62.4   &          \\
                                                                 &&       &    \textbf{20}   & 67.5 &        &  {\color{blue}{\textbf{65.5}}}   & 72.4   & 68.8   &           \\
        \cline{2-10}
        &\multirow{8}*{\makecell[c]{\textbf{tf-VAEGAN}}}& \multirow{4}*{\makecell[c]{\textbf{(G)UFSL}}}    &    \textbf{1}    & 82.4 & \multirow{4}*{8.97}  & 80.0  & 95.9  & 87.2 & \multirow{4}*{8.58}   \\
                                                              &&         &    \textbf{5}      & 92.1 &        &  {\color{red}{\textbf{90.7}}}  & 97.3   &  {\color{red}{\textbf{93.9}}}   &        \\
                                                               &&        &    \textbf{10}    & {\color{red}{\textbf{95.4}}} &        &  {\color{red}{\textbf{95.0}}}  & 97.4   &  {\color{red}{\textbf{96.2}}}  &          \\
                                                                &&        &    \textbf{20}    & 96.2 &        &  96.2   & 97.2   & 96.7   &           \\
       \cdashline{3-10}[1.5pt/2pt]         
       && \multirow{4}*{\makecell[c]{\textbf{(G)SFSL}}}      &    \textbf{1}     & 48.9 & \multirow{4}*{1.23}  &  {\color{blue}{\textbf{42.2}}} & 19.1  & 26.3  & \multirow{4}*{1.83}   \\
                                                              &&                      &    \textbf{5}     & 64.0 &        & 58.4   & 52.5   & 55.3   &        \\
                                                                &&                    &    \textbf{10}    & 67.0 &       & 61.6   & 67.6  &  64.5   &           \\
                                                                 &&                   &    \textbf{20}   & 68.7 &        & 64.1   & 74.4   &  {\color{blue}{\textbf{68.9}}}   &           \\
        \cline{2-10}
        &\multirow{8}*{\makecell[c]{\textbf{FREE}}}& \multirow{4}*{\makecell[c]{\textbf{(G)UFSL}}}    &    \textbf{1}    & 76.8 & \multirow{4}*{6.36}  & 74.5  & 93.1  & 82.8  & \multirow{4}*{7.25}    \\
                                                              &&         &    \textbf{5}    & 82.3 &        & 85.9   & 97.4   & 91.3   &          \\
                                                               &&        &    \textbf{10}    & 90.2 &        & 92.5   & 97.3   & 94.8   &         \\
                                                                &&        &    \textbf{20}    & 96.3 &        & 95.8   & 97.1   & 96.5   &         \\
       \cdashline{3-10}[1.5pt/2pt]       
       && \multirow{4}*{\makecell[c]{\textbf{(G)SFSL}}}                  &    \textbf{1}     & 48.9 & \multirow{4}*{6.31}  & 15.6  & 20.5  & 17.7 & \multirow{4}*{6.49}    \\
                                                               &&        &    \textbf{5}     & 63.1 &        &  {\color{blue}{\textbf{59.9}}}   & 45.2   & 51.5   &               \\
                                                                &&       &    \textbf{10}    & 67.0 &        & 62.2   & 62.4   & 62.3   &                  \\
                                                                 &&       &    \textbf{20}   &  {\color{blue}{\textbf{69.1}}} &        &  64.9   & 72.9    &  68.7      &   \\
        \cline{2-10}
        &\multirow{8}*{\makecell[c]{\textbf{GCM-CF}}}& \multirow{4}*{\makecell[c]{\textbf{(G)UFSL}}}    &    \textbf{1}    & 74.9 & \multirow{4}*{2.89}  & 73.9  & 90.6  & 81.4  & \multirow{4}*{2.91}    \\
                                                              &&         &    \textbf{5}    & 83.2 &        & 84.0   & 91.7   & 87.7   &       \\
                                                               &&        &    \textbf{10}    & 86.7 &        & 88.4   & 92.1   & 90.2   &           \\
                                                                &&        &    \textbf{20}    & 91.5 &        & 92.9   & 92.8   & 92.8   &         \\
       \cdashline{3-10}[1.5pt/2pt]         
       && \multirow{4}*{\makecell[c]{\textbf{(G)SFSL}}}                  &    \textbf{1}     & 41.8 & \multirow{4}*{0.54}  & 18.5  & 10.3  & 13.2 & \multirow{4}*{0.07}    \\
                                                               &&        &    \textbf{5}     & 54.7 &        & 42.5   & 44.2   & 43.3   &           \\
                                                                &&       &    \textbf{10}    & 60.6 &        & 41.0   & 55.4   & 47.1   &         \\
                                                                 &&       &    \textbf{20}   & 62.5 &        & 39.8   & 59.9   & 47.8   &         \\
        \Xhline{1pt}
        \multicolumn{10}{l}{{\makecell[l]{Note:  {\color{red}{\textbf{RED FONT}}} and {\color{blue}{\textbf{BLUE FONT}}} denote the best results for (G)UFSL and (G)SFSL, respectively}}}\\
       \vspace{-3em}
        \end{tabular}
\end{table}

\section{Conclusions}
In this paper, we systematically evaluate a significant number of state-of-the-art EAGMs, \ie, f-CLSWGAN, LisGAN, LsrGAN, CVAE, CADA-VAE, VAE-cFlow, f-VAEGAN-D2, tf-VAEGAN, FREE, and GCM, on five benchmark datasets, \ie, FLO, CUB, SUN, AWA2, and AWA, with a unified evaluation protocol for six any-shot scenarios, \ie, ZSL, GZSL, UFSL, GUFSL, SFSL, and GSFSL. Four types of visual features, \ie, the original features, naive features, finetuned features, and regularized features, and four types of semantic features, \ie, the original descriptions, naive descriptions, GRU-based descriptions, and imbalanced GRU-based descriptions, are designed and provided to explore the effects of features on EAGMs in depth.\par

First, for the first time, we systematically evaluate and prove that the current features used in benchmark datasets for ZSL and GZSL are somehow out-of-date by adequate experiments on a series of datasets and models. Our evaluation shows that accuracy improvement for ZSL and GZSL could be obtained from both the visual and semantic features. On the one hand, introducing the visual information and semantic information of seen classes into the visual features of EAGMs could improve model performance on both the seen and unseen classes. For example, the regularized features present an average $Z$ improvement of 6.2\% and 13.5\% on FLO and CUB, respectively, and present an average $H$ improvement of 8.7\% and 13.6\% on FLO and CUB, respectively. On the other hand, modifying the current semantic features could also boost EAGMs easily, \eg, applying GRU to release the overfitting of LSTM on seen classes and giving more weight to the loss of semantic features in model training. For FLO, the introduced imbalanced GRU-based descriptions present an average $Z$ improvement of 3.4\% in the ZSL setting and present an average $H$ improvement of 2.3\% in the GZSL setting. The significant improvements reveal the importance of features to EAGMs. The research on EAGM-based ZSL and GZSL should not be limited to the generator. The visual features and semantic features of EAGMs deserve demanding attention in future work.\par 

Second, different from previous reviews which usually used the results directly from their original papers, we first compare and reproduce ten typical EAGMs on five benchmark datasets with six any-shot scenarios. We summarize our works, including the four kinds of semantic and visual features, ten generative models, optimized parameters, and data splits for six tasks, in the publicly available GASL repository to foster the research on EAGMs. With GASL, the strong benchmark results of ZSL, GZSL, UFSL, GUFSL, SFSL, and GSFSL can be readily reproduced for various data insufficiency scenarios, and hence future methods, regularizations, and features can be sufficiently and fairly evaluated.

\vspace{-12 mm}
\begin{IEEEbiography}[{\includegraphics[width=1in,height=1.25in,clip,keepaspectratio]{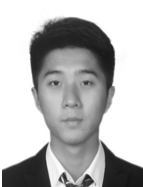}}]{Liangjun Feng}
received his B.Eng. from the North China Electric Power University, Beijing, China, in 2017. Now, he is pursuing his Ph.D. with the College of Control Science and Engineering, Zhejiang University, Hangzhou, China.  He has authored or co-authored more than 10 papers in peer-reviewed international journals. He is the reviewer of several journals, including IEEE TIP, IEEE TNNLS, and IEEE TII. His current research interests include zero-shot and few-shot learning.
\end{IEEEbiography}

\vspace{-12 mm}
\begin{IEEEbiography}[{\includegraphics[width=1in,height=1.25in,clip,keepaspectratio]{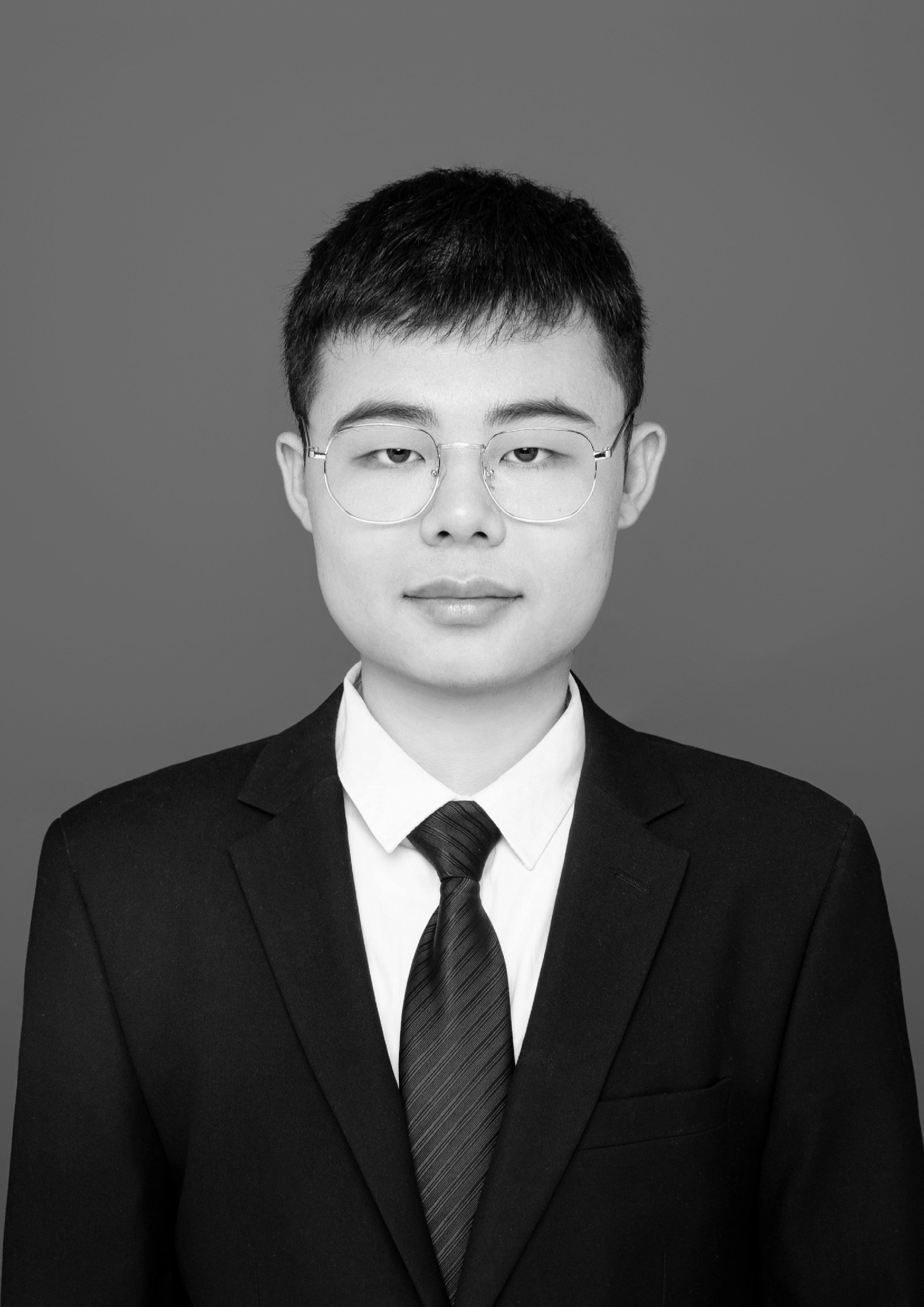}}]{Jiancheng Zhao}
received his B.Eng. from Zhejiang University, Zhejiang, China, in 2021. Now, he is pursuing his Ph.D. with the College of Control Science and Engineering, Zhejiang University, Hangzhou, China. His current research interests include zero-shot and few-shot learning.
\end{IEEEbiography}

\vspace{-12 mm}
\begin{IEEEbiography}[{\includegraphics[width=1in,height=1.25in,clip,keepaspectratio]{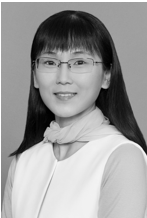}}]{Chunhui Zhao}
(SM’15) received the B.Eng., M.S., and Ph.D. degrees in Control Science and Engineering from the Department of Automation, Northeastern University, Shenyang, China, in 2003, 2006, and 2009, respectively. She was a Postdoctoral Fellow (January 2009-January 2012) at the Hong Kong University of Science and Technology, and the University of California, Santa Barbara, Los Angeles, CA, USA.\par
Chunhui Zhao is currently a full professor with the College of Control Science and Engineering, Zhejiang University, Hangzhou, China, since Jan. 2012. She has published over 170 papers in peer-reviewed international journals. Her research interests include machine learning and data mining. Prof. Zhao was a recipient of the National Science Fund for Distinguished Young Scholars, and New Century Excellent Talents in University, respectively.
\end{IEEEbiography}

\end{document}